\documentclass[lettersize,journal]{IEEEtran}
\usepackage{xcolor,soul,framed,caption} 

\usepackage[dvipsnames,svgnames,x11names]{xcolor}

\usepackage[pdftex]{graphicx}
\graphicspath{{../pdf/}{../jpeg/}}
\DeclareGraphicsExtensions{.pdf,.jpeg,.png}

\usepackage{enumitem}

\usepackage[cmex10]{amsmath}

\usepackage{array}
\usepackage{mdwmath}
\usepackage{mdwtab}
\usepackage{eqparbox}
\usepackage{url}
\usepackage{supertabular}                        \usepackage[colorlinks=true,
    linkcolor=blue,
    urlcolor=blue,
    citecolor=blue]{hyperref}    
\usepackage{amsmath,amssymb}
\usepackage{caption}
\usepackage{subcaption}
\usepackage{float}
\usepackage{booktabs}
\usepackage{multirow}
\usepackage{graphicx}
\usepackage{color}
\usepackage{tabularray}
\usepackage{hhline}
\usepackage{wasysym}
\usepackage[ruled, lined, longend, linesnumbered]{algorithm2e}
\usepackage{algorithmic}
\usepackage{tikz}
\usetikzlibrary{positioning,fit,shapes.misc,arrows.meta,calc}

\usepackage{tikz}
\usetikzlibrary{shapes,arrows.meta,positioning,fit,calc}

\usepackage{lipsum}

\usepackage{setspace}

\usepackage{titlesec}
\titlespacing*{\section}{0pt}{0.5\baselineskip}{0.3\baselineskip}
\titlespacing*{\subsection}{0pt}{0.4\baselineskip}{0.2\baselineskip}
\titlespacing*{\subsubsection}{0pt}{0.3\baselineskip}{0.1\baselineskip}

\makeatletter
\def\@IEEENORMtitlevspace{1\baselineskip}
\def\@IEEEMINtitlevspace{0.3\baselineskip}  
\makeatother


\begin{document}
\bstctlcite{IEEEexample:BSTcontrol}
    \title{AirFed: A Federated Graph-Enhanced Multi-Agent Reinforcement Learning Framework for Multi-UAV Cooperative Mobile Edge Computing}
  \author{Zhiyu Wang, Suman Raj, and Rajkumar Buyya
  \thanks{Zhiyu Wang, Suman Raj and Rajkumar Buyya are with the Quantum Cloud Computing and Distributed Systems (qCLOUDS) Laboratory, School of Computing and Information Systems, The University of Melbourne, Australia (e-mail: zhiywang1@student.unimelb.edu.au, suman.raj@student.unimelb.edu.au, rbuyya@unimelb.edu.au).}
  \thanks{Suman Raj is also with Department of Computational and Data Sciences, Indian Institute of Science, Bangalore, India (email: sumanraj@iisc.ac.in).}
}  

\maketitle
\IEEEaftertitletext{\vspace{-1.5em}}

\begin{abstract}
Multiple Unmanned Aerial Vehicles (UAVs) cooperative Mobile Edge Computing (MEC) systems face critical challenges in coordinating trajectory planning, task offloading, and resource allocation while ensuring Quality of Service (QoS) under dynamic and uncertain environments. Existing approaches suffer from limited scalability, slow convergence, and inefficient knowledge sharing among UAVs, particularly when handling large-scale IoT device deployments with stringent deadline constraints. This paper proposes AirFed, a novel federated graph-enhanced multi-agent reinforcement learning framework that addresses these challenges through three key innovations. First, we design dual-layer dynamic Graph Attention Networks (GATs) that explicitly model spatial-temporal dependencies among UAVs and IoT devices, capturing both service relationships and collaborative interactions within the network topology. Second, we develop a dual-Actor single-Critic architecture that jointly optimizes continuous trajectory control and discrete task offloading decisions. Third, we propose a reputation-based decentralized federated learning mechanism with gradient-sensitive adaptive quantization, enabling efficient and robust knowledge sharing across heterogeneous UAVs. Extensive experiments demonstrate that AirFed achieves 42.9\% reduction in weighted cost compared to state-of-the-art baselines, attains over 99\% deadline satisfaction and 94.2\% IoT device coverage rate, and reduces communication overhead by 54.5\%. Scalability analysis confirms robust performance across varying UAV numbers, IoT device densities, and system scales, validating AirFed's practical applicability for large-scale UAV-MEC deployments.
\end{abstract}


\IEEEpeerreviewmaketitle

\section{Introduction}

\IEEEPARstart{T}he Internet-of-Things (IoT) paradigm~\cite{gubbi2013internet} has fundamentally transformed data acquisition and decision-making across diverse spatio-temporal domains, including disaster management~\cite{dui2025multistage}, surveillance~\cite{9314091}, urban monitoring~\cite{lv2019infrastructure} and smart agriculture~\cite{vasisht2017farmbeats}. Despite this proliferation, contemporary IoT deployments deploy lightweight ground sensors and end-user devices with constrained computation, memory, and energy resources, thereby limiting their scalability to perform real-time analytics. The advent of 5G/6G communication technologies has facilitated large-scale data offloading from these devices to cloud or remote high-performance computing infrastructures for advanced processing~\cite{11074426}. Nevertheless, such centralized offloading becomes impractical for latency-sensitive applications due to variable end-to-end communication delays, and economically unsustainable as the monetary cost of cloud service usage escalates sharply with the increasing number of devices continuously generating data streams~\cite{wang2025tf}. Furthermore, in remote or disaster-affected regions where network connectivity is intermittent or bandwidth-constrained, cloud-based processing may not even be feasible. 

An emerging paradigm mitigating these challenges is Drones-as-a-Service (DaaS)~\cite{11169671}, which extends the principles of Mobile Edge Computing (MEC) into the aerial domain. Recent advances in embedded hardware have made it feasible to equip drones, or more formally Unmanned Aerial Vehicles (UAVs), with onboard accelerators such as the NVIDIA Jetson Orin Nano, featuring 1024 CUDA cores, a 6-core Arm Cortex-A78AE CPU, and 8 GB of unified memory, all within a compact 100 × 79 mm form factor, a power envelope of 7–15 W and costing just around US$\$400$~\cite{orin-tech-specs}. Within this paradigm, UAVs function as mobile sensing and compute nodes, dynamically repositioned to provide localized edge computing services to ground IoT devices. They can operate either autonomously or as part of a cooperative aerial fleet, forming a distributed edge computing layer serving ground IoT devices. This enables efficient execution of Deep Neural Network (DNN) inference tasks on-board, facilitating real-time analytics without dependence on continuous backhaul connectivity~\cite{shen2023survey}.

When integrated into a DaaS orchestration framework, UAVs can form a collaborative service fabric in which incoming requests from ground IoT devices are dynamically mapped to aerial nodes based on proximity, residual compute capacity, and deadline constraints~\cite{10795214}. A UAV that receives a service request may either process it locally or offload it to another UAV within the fleet, similar to a distributed microservice invocation, thus ensuring Quality-of-Service (QoS) guarantees and efficient resource utilization. 

\subsection{Challenges}
Realizing effective UAV-enabled MEC in IoT environments entails addressing several fundamental challenges arising from the distributed, dynamic, and resource-constrained nature of aerial edge computing. UAV fleets typically comprise heterogeneous platforms with varying computational capabilities, where high-end UAVs equipped with accelerators can execute inference tasks within milliseconds, while lower-end UAVs may require several seconds for the same workload \cite{zeng2023a3d}. This heterogeneity necessitates intelligent task scheduling that efficiently exploits resource diversity while managing the critical energy trade-offs among onboard computation, wireless communication, and flight operations under stringent battery constraints.

Further, the inherent mobility of both UAVs and ground IoT devices, coupled with stochastic task arrivals and time-varying wireless channel conditions, introduces significant uncertainty into the system. Traditional approaches that rely on accurate system models and deterministic parameters struggle to maintain performance under such dynamic conditions, as environmental changes necessitate frequent re-optimization with high computational overhead \cite{wang2024deep}. Real-time decision-making in this context requires adaptive policies capable of responding to evolving network states without complete system re-design.

Moreover, multi-UAV systems require distributed coordination for collaborative task allocation and resource sharing, yet centralized control architectures introduce single-point failure risks and communication bottlenecks for state aggregation \cite{liao2023decentralized}. Achieving efficient decentralized coordination across heterogeneous UAVs while avoiding resource overload remains a fundamental challenge, particularly when UAVs must cooperatively serve overlapping areas and relay tasks among themselves.

Finally, real-world IoT applications impose stringent QoS requirements, demanding both timely task completion within application-specific deadlines and adequate coverage for distributed IoT devices \cite{centenaro2021survey}. Jointly optimizing task scheduling and UAV trajectory planning to satisfy these QoS requirements under dynamic conditions is non-trivial, especially when considering the cascading effects of UAV mobility on network topology and task execution latency.




\subsection{Contributions}
This paper proposes AirFed, a federated graph-enhanced multi-agent reinforcement learning framework for multi-UAV cooperative mobile edge computing. AirFed achieves joint optimization of UAV trajectory planning and task offloading through spatial-temporal graph modeling, Constrained Multi-Agent Reinforcement Learning (CMARL), and communication-efficient decentralized federated learning, while guaranteeing coverage and deadline QoS requirements. The main contributions of our paper are as follows:

\begin{enumerate}[leftmargin=*]
\item We establish a comprehensive system model encompassing multi-hop task offloading, UAV heterogeneity, coverage guarantees, and deadline constraints, and formulate the joint optimization as a multi-objective Mixed-integer non-linear programming (MINLP) problem.
    
\item We design dual-layer dynamic Graph Attention Network (GATs) with Gated Recurrent Unit (GRU) to model spatial-temporal dependencies, and develop a dual-Actor single-Critic architecture that unifies continuous trajectory control and discrete task offloading within a hybrid action space.
    
\item We propose reputation-based decentralized federated learning with gradient-sensitive adaptive quantization, enabling efficient knowledge sharing across UAVs while reducing communication overhead.
    
\item We conduct comprehensive experiments demonstrating AirFed's superior performance in convergence, QoS guarantees, and communication efficiency, with ablation and scalability analysis validating effectiveness across different system scales.
\end{enumerate}

The rest of the paper is organized as follows: Section~\ref{related_work} reviews related work and identifies research gaps; Section~\ref{system_model} establishes the system model and formulates the optimization problem; Section~\ref{sec:airfed} presents the detailed design of the AirFed framework; Section~\ref{evaluation} evaluates the performance of the framework through extensive experiments; Section~\ref{conclusions} concludes the paper and discusses future work.

\section{Related Work}
\label{related_work}
Recent research on UAV-assisted MEC has focused primarily on joint optimization of trajectory planning and resource allocation to improve system performance. Existing approaches can be broadly categorized into optimization-based methods and learning-based methods.

\subsection{Optimization-based Approaches}
Pervez et al. \cite{10179258} proposed a Block Coordinate Descent (BCD) method combined with game theory and Successive Convex Approximation (SCA) for multi-UAV assisted MEC networks. The approach jointly optimizes task offloading decisions and UAV trajectory to minimize energy and latency-based cost function. Qi et al. \cite{10130436} proposed an SCA-based joint optimization algorithm for UAV-relaying-assisted MEC networks with moving users. The approach minimizes average task completion time by optimizing communication bandwidth, CPU frequency, task division ratio, and UAV three-dimensional location deployment. Dai et al. \cite{10077418} proposed a Lyapunov optimization-based method combined with Markov chain approximation for online UAV-assisted task offloading in vehicular edge computing networks. The approach minimizes vehicular task delay under long-term UAV energy constraints without requiring future information. He et al. \cite{10621306} proposed an online joint optimization approach based on Lyapunov method for QoE maximization in UAV-enabled MEC. The method employs a two-stage optimization combining game theory and convex optimization to solve per-slot task offloading, resource allocation, and UAV trajectory planning problems. Xu et al. \cite{10654497} proposed a surrogate Lagrangian relaxation method with hybrid numerical techniques for service selection in MEC-based UAV last-mile delivery systems. The approach addresses heterogeneous service requirements by jointly optimizing delivery and computational service selection to minimize UAV energy consumption and service response time. Tun et al. \cite{10795214} proposed a BCD approach for joint UAV deployment and resource allocation in MEC-enabled integrated space-air-ground networks. The method decomposes the problem into four subproblems solved by matching game, concave-convex procedure, SCA, and block successive upper-bound minimization approaches to minimize device and UAV energy consumption. Du et al. \cite{10857309} proposed an online optimization framework based on Lyapunov optimization and surrogate Lagrangian relaxation for hierarchical collaborative MEC systems. The approach decouples decisions across time slots and employs hybrid numerical techniques for service placement, task scheduling, and resource allocation to minimize energy consumption while ensuring service placement stability. Gao et al. \cite{gao2025improving} proposed a BCD-based method with SCA technique for multi-UAV assisted MEC to minimize task completion time. The approach iteratively optimizes UAV-user association, UAV trajectory planning, and transmit power allocation to improve quality of experience.

\subsection{Learning-based Approaches}
Ning et al. \cite{ning2024multi} proposed MUTO, a Multi-Agent Deep Reinforcement Learning (MADRL) algorithm with prioritized experience replay for UAV trajectory optimization in differentiated services scenarios. The approach formulates a Markov game model and employs Actor-Critic architecture to achieve distributed trajectory control of multiple UAVs while minimizing energy consumption based on local observations. Song et al. \cite{song2024aoi} proposed a multi-objective learning approach based on Proximal Policy Optimization (PPO) combined with genetic operators for aerial-ground collaborative MEC. The method addresses latency and energy tradeoff by optimizing UAV flight paths and task offloading ratios through crossover and mutation operations at policy network parameter level. Li et al. \cite{li2024computation} proposed an improved federated Deep Deterministic Policy Gradient (IF-DDPG) algorithm for computation offloading in multi-UAV assisted MEC. The method enhances traditional federated DDPG through dual experience replay and mixed noise exploration to minimize task response delay and user energy consumption. Chen et al. \cite{chen2025joint} proposed the JTORA algorithm integrating Soft Actor-Critic (SAC) and Lyapunov optimization techniques for joint trajectory optimization and resource allocation in UAV-MEC systems. The approach employs Lyapunov techniques for problem transformation and combines DRL with convex optimization to minimize mobile user energy consumption. 

\renewcommand{\arraystretch}{1.5}
\begin{table*}[t]
\centering
\caption{A qualitative comparison of AirFed with existing related works}
\label{tab:related_works}
\resizebox{\textwidth}{!}{%
\begin{tabular}{|c|c|c|c|c|c|c|c|c|c|c|c|c|c|c|}
\hline
\multirow{3}{*}{\textbf{Work}} & \multicolumn{4}{c|}{\textbf{Problem Formulation}} & \multicolumn{3}{c|}{\textbf{System Modeling}} & \multicolumn{4}{c|}{\textbf{Solution Approach}} & \multicolumn{3}{c|}{\textbf{Performance Evaluation}} \\ \cline{2-15}
 & \multicolumn{3}{c|}{\textbf{Optimization Objective}} & \multirow{2}{*}{\begin{tabular}[c]{@{}c@{}}\textbf{Decision}\\\textbf{Variables}\end{tabular}} & \multirow{2}{*}{\begin{tabular}[c]{@{}c@{}}\textbf{Multi-hop}\\\textbf{Offloading}\end{tabular}} & \multirow{2}{*}{\begin{tabular}[c]{@{}c@{}}\textbf{UAV}\\\textbf{Heterogeneity}\end{tabular}} & \multirow{2}{*}{\begin{tabular}[c]{@{}c@{}}\textbf{Spatial-Temporal}\\\textbf{Modeling}\end{tabular}} & \multirow{2}{*}{\begin{tabular}[c]{@{}c@{}}\textbf{Solution}\\\textbf{Method}\end{tabular}} & \multirow{2}{*}{\begin{tabular}[c]{@{}c@{}}\textbf{Cooperation}\\\textbf{Mechanism}\end{tabular}} & \multirow{2}{*}{\begin{tabular}[c]{@{}c@{}}\textbf{Knowledge}\\\textbf{Transfer}\end{tabular}} & \multirow{2}{*}{\begin{tabular}[c]{@{}c@{}}\textbf{Communication}\\\textbf{Efficiency}\end{tabular}} & \multicolumn{2}{c|}{\textbf{QoS Guarantee}} & \multirow{2}{*}{\textbf{Scalability}} \\ \cline{2-4} \cline{13-14}
 & \textbf{Time} & \textbf{Energy} & \textbf{Multi-Obj} &  &  &  &  &  &  &  &  & \textbf{Coverage} & \textbf{Deadline} &  \\ \hline
Pervez et al. \cite{10179258} & \checkmark & \checkmark & \checkmark & Joint & $\times$ & \checkmark & None & Optimization & Centralized & None & None & $\times$ & $\times$ & \checkmark \\ \hline
Qi et al. \cite{10130436} & \checkmark & $\times$ & $\times$ & Joint & $\times$ & \checkmark & None & Optimization & Centralized & None & None & $\times$ & $\times$ & \checkmark \\ \hline
Dai et al. \cite{10077418} & \checkmark & $\times$ & $\times$ & Offloading & $\times$ & \checkmark & None & Optimization & Centralized & None & None & $\times$ & $\times$ & \checkmark \\ \hline
He et al. \cite{10621306} & \checkmark & \checkmark & \checkmark & Joint & $\times$ & \checkmark & None & Optimization & Centralized & None & None & $\times$ & \checkmark & $\times$ \\ \hline
Xu et al. \cite{10654497} & \checkmark & \checkmark & \checkmark & Joint & $\times$ & \checkmark & None & Optimization & Centralized & None & None & $\times$ & $\times$ & \checkmark \\ \hline
Tun et al. \cite{10795214} & $\times$ & \checkmark & $\times$ & Joint & \checkmark & \checkmark & None & Optimization & Centralized & None & None & $\times$ & \checkmark & \checkmark \\ \hline
Du et al. \cite{10857309} & $\times$ & \checkmark & $\times$ & Offloading & $\times$ & \checkmark & None & Optimization & Centralized & None & None & $\times$ & \checkmark & \checkmark \\ \hline
Gao et al. \cite{gao2025improving} & \checkmark & $\times$ & $\times$ & Joint & $\times$ & \checkmark & None & Optimization & Centralized & None & None & $\times$ & $\times$ & \checkmark \\ \hline
Ning et al. \cite{ning2024multi} & $\times$ & \checkmark & $\times$ & Trajectory & $\times$ & \checkmark & None & MARL & Decentralized & Experience & None & $\times$ & $\times$ & \checkmark \\ \hline
Song et al. \cite{song2024aoi} & \checkmark & \checkmark & \checkmark & Joint & $\times$ & \checkmark & None & Single DRL & None & None & None & $\times$ & $\times$ & \checkmark \\ \hline
Li et al. \cite{li2024computation} & \checkmark & \checkmark & \checkmark & Joint & $\times$ & \checkmark & None & Single DRL & Centralized & Federated & None & $\times$ & \checkmark & \checkmark \\ \hline
Chen et al. \cite{chen2025joint} & $\times$ & \checkmark & $\times$ & Joint & $\times$ & $\times$ & None & Single DRL & None & None & None & $\times$ & $\times$ & \checkmark \\ \hline
\textbf{AirFed (Ours)} & \checkmark & \checkmark & \checkmark & Joint & \checkmark & \checkmark & Dual-layer GATs & MARL & Decentralized & Federated & Adaptive Quantization & \checkmark & \checkmark & \checkmark \\ \hline
\end{tabular}%
}
\end{table*}
\renewcommand{\arraystretch}{1}

\subsection{A Qualitative Comparison}
To systematically position our work and identify research gaps, we conduct a comprehensive qualitative comparison of related works presented in Table~\ref{tab:related_works}.

\subsubsection{Comparative Analysis Dimensions}
We evaluate related works across four dimensions: problem formulation, system properties, solution approach, and performance guarantees.

\textbf{Problem Formulation} dimensions assess the optimization objectives and decision variables. \textbf{Optimization Objective} includes three sub-dimensions: Time (whether latency is optimized), Energy (whether energy consumption is optimized), and Multi-Objective (whether multiple objectives are jointly optimized). \textbf{Decision Variables} categorizes control variables: trajectory planning only, task offloading only, or joint optimization of both.

\textbf{System Modeling} dimensions characterize how the UAV-MEC network is represented. \textbf{Multi-hop Offloading} indicates whether the system supports task forwarding among UAVs. \textbf{UAV Heterogeneity} specifies whether UAVs are heterogeneous or homogeneous in terms of computational and communication capabilities. \textbf{Spatial-Temporal Modeling} indicates whether the approach explicitly models the spatio-temporal relationships in the UAV-MEC network.

\textbf{Solution Approach} dimensions analyze the algorithmic paradigm and technical architecture. \textbf{Solution Method} reflects the solution paradigm: traditional optimization techniques, single-agent DRL, or MADRL. \textbf{Cooperation Mechanism} describes how multiple UAVs coordinate: no cooperation, centralized coordination (requiring a central controller), or decentralized peer-to-peer collaboration. \textbf{Knowledge Transfer} evaluates how learning knowledge is shared among agents: no sharing, experience sharing, or model sharing (e.g., federated learning). \textbf{Communication Efficiency} examines whether communication overhead optimization techniques are employed.

\textbf{Performance Evaluation} dimensions assess the system's performance characteristics and capabilities. \textbf{QoS Guarantee} includes two sub-dimensions: \textbf{Coverage Guarantee} indicates whether IoT device coverage requirements are ensured, and \textbf{Deadline Guarantee} indicates whether task deadline constraints are ensured. \textbf{Scalability} assesses whether the approach evaluates its capability to handle system scale growth (increasing number of UAVs and devices).

\subsubsection{Research Gap Identification}
Based on the systematic analysis presented in Table~\ref{tab:related_works}, we identify five critical research gaps in existing UAV-MEC solutions:

\textbf{Gap 1 - Inadequate Multi-hop Offloading Support and Spatio-Temporal Modeling.}
First, multi-hop task offloading remains largely unexplored, with only Tun et al.~\cite{10795214} addressing this scenario through traditional optimization. However, in practical deployments, direct communication between users and optimal serving UAVs may be infeasible due to coverage limitations, severe path loss, or resource contention \cite{ning2023mobile}. Multi-hop relay through intermediate UAVs is essential for extending service coverage and enabling flexible load balancing. Second, the table reveals that all existing works report \textit{None} for spatial-temporal modeling. UAV-MEC networks inherently form dynamic graphs where nodes (UAVs) and edges (wireless links) evolve over time with complex spatial dependencies \cite{zhou2020mobile}. However, current approaches treat UAVs as independent entities or employ flat vector representations, failing to capture network topology and spatio-temporal evolution patterns.

\textbf{Gap 2 - Dominance of Centralized Optimization with Limited Adaptability.}
Table~\ref{tab:related_works} shows that 67\% of existing works (8 out of 12) employ traditional optimization methods, and all optimization-based works adopt centralized cooperation mechanisms, requiring a central controller to collect global state and solve complex MINLP problems. This centralized architecture introduces single-point failure risks and communication bottlenecks for state aggregation. More fundamentally, optimization methods rely on accurate system models and deterministic parameter assumptions, while real systems face stochastic task arrivals, time-varying channel conditions, and unpredictable interference patterns \cite{zhou2024predictable}. When environments deviate from modeling assumptions, these methods require complete re-optimization with high computational complexity, making online adaptation infeasible and unable to cope with the dynamics and uncertainties of UAV-MEC networks.

\textbf{Gap 3 - Limited Decentralized Cooperation with Efficient Knowledge Sharing.}
A critical gap exists in achieving decentralized cooperation with efficient knowledge transfer. Li et al.~\cite{li2024computation} employ federated learning for model-level aggregation but remain centralized in cooperation, requiring a central controller for decision-making and creating single-point failure risks. Conversely, Ning et al.~\cite{ning2024multi} achieve decentralized cooperation but rely on experience sharing, which lacks the abstraction benefits of model-level knowledge and exhibits lower sample efficiency compared to federated aggregation. Table~\ref{tab:related_works} reveals that no existing work combines decentralized cooperation with federated learning, preventing systems from simultaneously achieving distributed autonomy and efficient model-level knowledge transfer. This gap is particularly critical for UAV networks requiring both operational independence and rapid collaborative learning.

\textbf{Gap 4 - Systematic Neglect of Communication Efficiency.}
The table shows that all existing works report \textit{None} for communication efficiency, reflecting an implicit assumption that communication costs are negligible. However, UAV-MEC networks already face significant communication demands for task offloading and result delivery. Optimization-based and learning-based methods introduce additional communication overhead for coordination, state sharing, or model synchronization among UAVs. Given that UAV wireless links typically provide limited throughput and are severely energy-constrained \cite{ning2023mobile}, without explicit communication-aware mechanisms, the aggregate communication burden from both system operations and algorithmic overhead may fundamentally constrain the practical viability of these solutions in resource-limited UAV deployments.

\textbf{Gap 5 - Incomplete Quality-of-Service Guarantee Mechanisms.}
The table reveals incomplete treatment of QoS guarantees in two critical dimensions. First, no existing work considers coverage guarantee, which ensures users remain within UAV service range. Existing works implicitly assume static user-UAV associations or focus solely on optimizing performance for already-connected users, neglecting the challenge of maintaining continuous coverage in dynamic scenarios. Second, only four works (He et al.~\cite{10621306}, Tun et al.~\cite{10795214}, Du et al.~\cite{10857309}, Li et al.~\cite{li2024computation}) incorporate deadline constraints to guarantee timely task completion. No work achieves joint coverage-deadline guarantees, while real-world applications often demand both service availability (coverage) and responsiveness (deadline satisfaction), limiting the applicability of existing solutions to scenarios with comprehensive QoS requirements.

To address these gaps, the proposed AirFed framework provides systematic solutions. For Gap 1, AirFed supports multi-hop task offloading and employs Dual-layer GATs to explicitly model spatial-temporal relationships in UAV-MEC networks. For Gap 2, it adopts CMARL to replace traditional optimization, enabling online adaptive decision-making under system constraints through data-driven approaches. For Gap 3, it is the first to organically combine decentralized cooperation with federated learning, enabling each UAV to make autonomous decisions while performing efficient model-level knowledge sharing. For Gap 4, it integrates a gradient-sensitive adaptive quantization mechanism to significantly reduce communication overhead, ensuring deployability in bandwidth-limited UAV-MEC networks. For Gap 5, it simultaneously provides coverage and deadline guarantees through coordinated UAV trajectory planning and multi-hop task offloading. As shown in Table~\ref{tab:related_works}, AirFed is the only work that provides comprehensive solutions across all critical dimensions.

\section{System Model and Problem Formulation}
\label{system_model}
\begin{figure*}[h]
\centering
\includegraphics[scale=0.4]{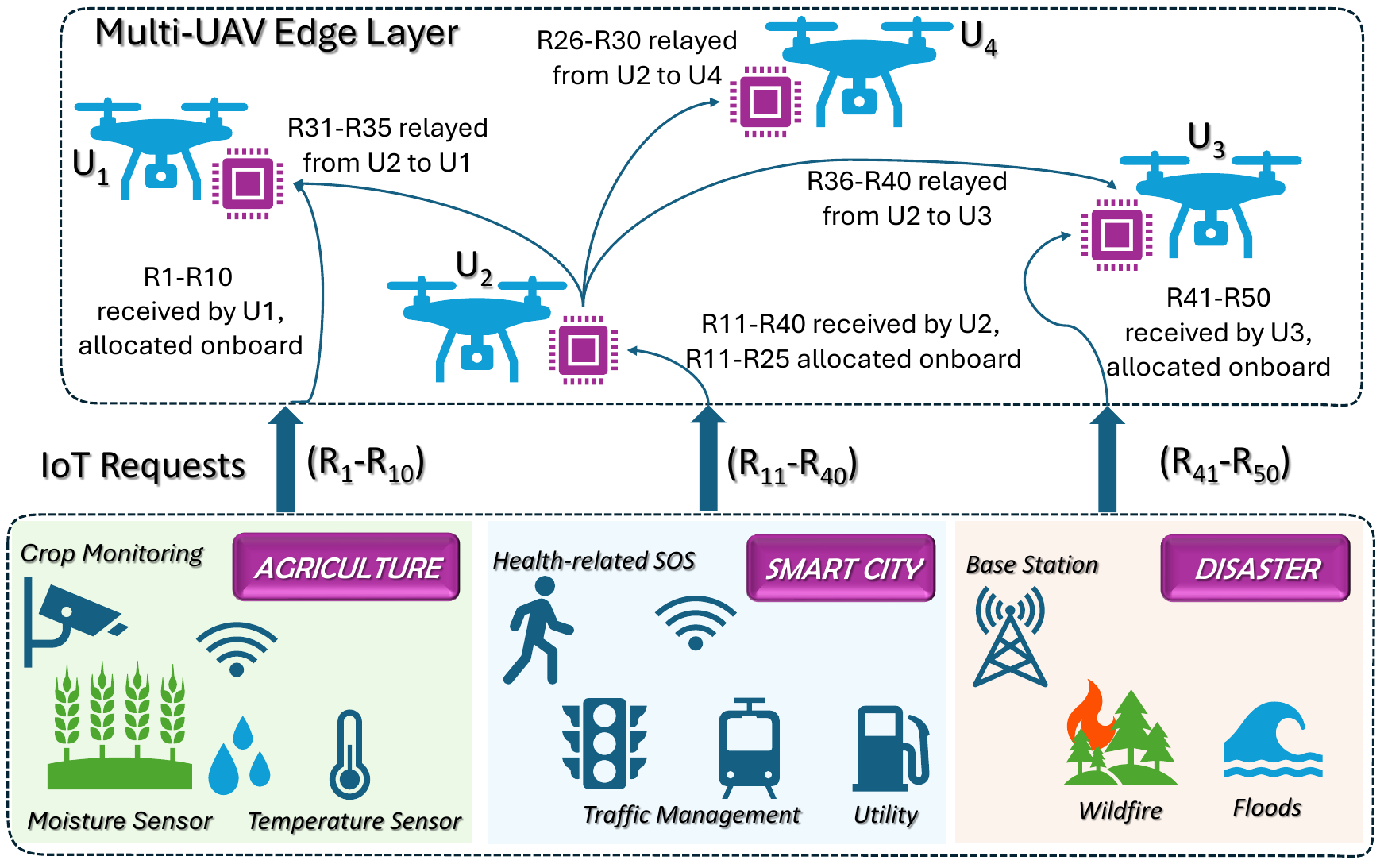}
\caption{System model showing multi-UAV cooperative edge computing serving 50 requests across three application domains with heterogeneous spatial distributions.}
\label{fig:system_model}
\end{figure*}

This section establishes the mathematical model for the multi-UAV cooperative mobile edge computing system and formulates the optimization problem. Fig.~\ref{fig:system_model} illustrates a representative deployment where four UAVs serve 50 computation requests from multiple IoT devices across agriculture, smart city, and disaster response domains.

\subsection{Network and Spatial Model}
We consider a multi-UAV cooperative system that provides mobile edge computing services to distributed IoT devices. This section establishes the mathematical model for the network topology and spatial relationships.

The system comprises $K$ UAVs, denoted as $\mathcal{U} = \{u_k \mid 1 \leq k \leq K\}$. The state of each UAV $u_k$ at time $t$ can be represented by a four-tuple:
\begin{align}
s_k(t) = \langle pos_k(t), v_k(t), energy_k(t), load_k(t) \rangle,
\end{align}
where $pos_k(t) = (x_k(t), y_k(t), h_k)$ denotes the three-dimensional coordinate position, $v_k(t) = (v_{x,k}(t), v_{y,k}(t))$ is the current velocity vector, $energy_k(t)$ is the remaining battery level, and $load_k(t)$ is the current computational load.

The service area contains $M$ IoT devices distributed across the region, denoted as $\mathcal{D} = \{d_m \mid 1 \leq m \leq M\}$. Each IoT device $d_m$ has a fixed geographical location $loc_m = (x_m, y_m)$. The task generation rate of device $d_m$ follows $\lambda_m(t) \sim \text{Poisson}(\mu_m)$, where $\mu_m$ is the parameter of the average task generation rate of device $d_m$, reflecting the random arrival characteristics of computational tasks.

The communication coverage capability of UAV $u_k$ over IoT device $d_m$ depends on their spatial distance. The Euclidean distance between the device and UAV is:
\begin{align}
d_{m,k}(t) = \sqrt{(x_k(t) - x_m)^2 + (y_k(t) - y_m)^2 + h_k^2}.
\end{align}
Based on the free-space path loss model \cite{goldsmith2005wireless}, the channel gain between the device and UAV is:
\begin{align}
|h_{m,k}|^2 = \frac{G_0}{d_{m,k}(t)^2},
\end{align}
where $G_0$ is the gain constant of the reference channel, reflecting the basic characteristics of the signal propagation environment. The Received Signal Strength Indicator (RSSI) at IoT device $d_m$ from UAV $u_k$ is:
\begin{align}
RSSI_{k,m}(t) = P_k^{tx} \times |h_{m,k}|^2 = \frac{P_k^{tx} \cdot G_0}{d_{m,k}(t)^2},
\end{align}
where $P_k^{tx}$ is the transmission power of UAV $u_k$.

When the signal strength exceeds the minimum reception threshold $RSSI_{min}$, the UAV is considered capable of providing effective service to that device. Therefore, the coverage status of UAV $u_k$ over device $d_m$ at time $t$ can be expressed using an indicator function:
\begin{align}
I_{k,m}^{cov}(t) = \begin{cases}
1, & \text{if } RSSI_{k,m}(t) \geq RSSI_{min} \\
0, & \text{otherwise}
\end{cases}.
\end{align}

\subsection{Task Completion Time Model}
In UAV-assisted mobile edge computing systems, task completion time is influenced by multiple factors including wireless channel conditions, UAV computational capabilities, network topology, and offloading decision strategies. This section establishes a path-based task completion time model to quantify task processing delay.

Each computational task generated by the IoT device $d_m$ is represented by a four-tuple $\tau_l = \langle w_l, s_l, d_l, c_l \rangle$, where $w_l$ denotes the required CPU cycles, $s_l$ and $d_l$ represent input and output data sizes, respectively, and $c_l$ indicates the deadline constraint.

Considering the limited computational capabilities and simple network access mechanisms of IoT devices, devices employ a signal strength-based UAV selection strategy. The device $d_m$ selects the UAV $u_{k^*}$ that provides the strongest signal coverage as its service node:
\begin{align}
u_{k^*} = \arg\max_{u_k \in \mathcal{U}} RSSI_{k,m}(t).
\end{align}

When a UAV receives a task request, it must decide whether to execute the task locally or forward it to other UAVs for processing. Each task follows a path model where the processing path for task $\tau_l$ is represented as an ordered sequence $\mathcal{P}_l = \{u_{k_1}, u_{k_2}, ..., u_{k_h}\}$, where $u_{k_1}$ is the serving UAV that receives the task, $u_{k_h}$ is the final UAV that executes the computation, and $h \geq 1$ is the path length. When $h=1$, it represents local processing; when $h>1$, it represents multi-hop forwarding processing. The end-to-end task completion time comprises three main stages:
\begin{align}
T_l^{total} = T_l^{uplink} + T_l^{path} + T_l^{downlink}.
\end{align}

The uplink transmission time $T_l^{uplink}$ represents the time required for the IoT device to transmit task data to the serving UAV:
\begin{align}
T_l^{uplink} = \frac{s_l}{R_{m,k_1}^{up}}.
\end{align}
Based on the Shannon-Hartley theorem\cite{goldsmith2005wireless}, the uplink channel capacity $R_{m,k_1}^{up}$ is:
\begin{align}
\scalebox{1}{$R_{m,k_1}^{up} = B \log_2\left(1 + \frac{P_m^{tx} \cdot |h_{m,k_1}|^2}{N_0 + \sum_{m' \neq m} P_{m'}^{tx} \cdot |h_{m',k_1}|^2}\right)$},
\end{align}
where $B$ is the channel bandwidth, $P_m^{tx}$ is the IoT device transmission power, $N_0$ is the noise power, and the summation term $\sum_{m' \neq m} P_{m'}^{tx} \cdot |h_{m',k_1}|^2$ represents the interference power from other simultaneously transmitting devices. 

The path processing time $T_l^{path}$ can be decomposed into four distinct components:
\begin{align}
\scalebox{1}{$T_l^{path} = T_l^{decision} + T_l^{forward} + T_l^{process} + T_l^{return}$},
\end{align}
where each component represents a specific stage of the processing pipeline.

The cumulative decision time $T_l^{decision}$ accounts for the offloading decisions (local processing or forwarding) made by each UAV along the path:
\begin{align}
T_l^{decision} = \sum_{i=1}^{h} T_{k_i}^{decision}.
\end{align}

The forwarding transmission time $T_l^{forward}$ represents the inter-UAV communication time for task data:
\begin{align}
T_l^{forward} = \sum_{i=1}^{h-1} \frac{s_l}{R_{k_i,k_{i+1}}^{inter}},
\end{align}
where the inter-UAV communication capacity $R_{k,k'}^{inter}$ is modeled based on the line-of-sight propagation characteristics \cite{zeng2023three} of air-to-air links:
\begin{align}
R_{k,k'}^{inter} = B_{inter} \log_2\left(1 + \frac{P_{k}^{tx} \cdot |h_{k,k'}|^2}{N_0}\right),
\end{align}
where $B_{inter}$ is the inter-UAV communication bandwidth, typically much larger than the air-to-ground communication bandwidth $B$, and $P_{k}^{tx}$ is the transmission power of UAV $u_k$. $|h_{k,k'}|^2$ is the inter-UAV channel gain, modeled as:
\begin{align}
|h_{k,k'}|^2 = \frac{G_{inter}}{d_{k,k'}^2},
\end{align}
where $d_{k,k'}$ is the Euclidean distance between two UAVs, and $G_{inter}$ is the reference gain constant for inter-UAV communication.

The processing time $T_l^{process}$ at the final UAV includes both queuing delay and computation execution:
\begin{align}
T_l^{process} = T_{k_h}^{queue} + \frac{w_l}{f_{k_h}},
\end{align}
where $f_{k_h}$ represents the computational frequency of the executing UAV.

The result return time $T_l^{return}$ covers the transmission of computation results back along the reverse path:
\begin{align}
T_l^{return} = \sum_{i=h}^{2} \frac{d_l}{R_{k_i,k_{i-1}}^{inter}}.
\end{align}

The downlink transmission time $T_l^{downlink}$ represents the time for transmitting computation results from the serving UAV back to the original IoT device:
\begin{align}
T_l^{downlink} = \frac{d_l}{R_{k_1,m}^{down}}.
\end{align}
The downlink channel capacity $R_{k_1,m}^{down}$ is modeled similarly to the uplink, but UAVs typically have higher transmission power than IoT devices.

\subsection{Energy Consumption Model}
The energy consumption in UAV-assisted mobile edge computing systems primarily stems from the multi-dimensional operational overhead of UAVs. To accurately reflect the different impacts of trajectory optimization and task processing decisions, this section establishes a UAV energy consumption model.

The total energy consumption of a UAV comprises two independent components: trajectory flight energy consumption and task processing energy consumption:
\begin{align}
E_k^{total} = E_k^{trajectory} + \sum_{l \in \mathcal{L}_k} E_{k,l}^{task},
\end{align}
where $\mathcal{L}_k$ represents the set of all tasks processed by UAV $u_k$.

The trajectory flight energy consumption $E_k^{trajectory}$ reflects the total energy consumption of UAVs executing mobility strategies. UAV flight energy consumption depends on the instantaneous velocity, with higher velocities resulting in increased aerodynamic drag and power requirements. The instantaneous flight power is modeled based on aerodynamic characteristics:
\begin{align}
\label{eq:p_flight}
P_k^{flight}(||v_k(t)||_2) = P_k^{hover} + \frac{1}{2} \rho A C_D ||v_k(t)||_2^3,
\end{align}
where $P_k^{hover}$ is the hovering power, $\rho$ is the air density, $A$ is the effective drag area, $C_D$ is the drag coefficient, and $||v_k(t)||_2$ is the flight speed. The trajectory energy consumption is calculated by integrating power consumption over the flight duration:
\begin{align}
E_k^{trajectory} = \int_0^{T_k^{flight}} P_k^{flight}(||v_k(t)||_2) dt
\end{align}

The task processing energy consumption is based on the three-stage structure:
\begin{align}
E_{k,l}^{task} = E_{k,l}^{uplink} + E_{k,l}^{path} + E_{k,l}^{downlink}.
\end{align}

The uplink communication energy consumption $E_{k,l}^{uplink}$ represents the energy consumption for the serving UAV to receive task data from IoT devices:
\begin{align}
E_{k,l}^{uplink} = P_k^{rx} \cdot T_l^{uplink},
\end{align}
where $P_k^{rx}$ is the air-to-ground communication receiving power of the UAV.

The path processing energy consumption $E_{k,l}^{path}$ can be decomposed into four components:
\begin{align}
E_{k,l}^{path} = E_{k,l}^{decision} + E_{k,l}^{forward} + E_{k,l}^{process} + E_{k,l}^{return}.
\end{align}

The decision energy consumption $E_{k,l}^{decision}$ represents the cumulative energy consumption of all UAVs along the path executing offloading decision algorithms:
\begin{align}
E_{k,l}^{decision} = \sum_{i=1}^{h} P_{k_i}^{cpu} \cdot T_{k_i}^{decision},
\end{align}
where $P_{k_i}^{cpu}$ is the CPU power consumption of UAV $u_{k_i}$ when executing decision algorithms.

The forwarding communication energy consumption $E_{k,l}^{forward}$ represents the energy consumption for inter-UAV task data transmission, including both transmission and reception components:
\begin{align}
\scalebox{0.9}{$E_{k,l}^{forward} = \sum_{i=1}^{h-1} \left( P_{k_i}^{tx} \cdot \frac{s_l}{R_{k_i,k_{i+1}}^{inter}} + P_{k_{i+1}}^{rx} \cdot \frac{s_l}{R_{k_i,k_{i+1}}^{inter}} \right)$},
\end{align}
where $P_{k_i}^{tx}$ is the inter-UAV communication transmission power of sending UAV $u_{k_i}$, and $P_{k_{i+1}}^{rx}$ is the inter-UAV communication reception power of receiving UAV $u_{k_{i+1}}$.

The processing energy consumption $E_{k,l}^{process}$ occurs at the final executing UAV, including the basic power consumption during queuing delay and actual computation energy consumption:
\begin{align}
E_{k,l}^{process} = P_{k_h}^{idle} \cdot T_{k_h}^{queue} + E_{k_h,l}^{compute}.
\end{align}
The computation energy consumption $E_{k_h,l}^{compute}$ is modeled based on the dynamic power characteristics of CMOS digital circuits. According to the CMOS power formula \cite{rabaey2002digital}, the dynamic power consumption is given by $P = \alpha C V^2 f$, where $\alpha$ is the activity factor, $C$ is the load capacitance, $V$ is the supply voltage, and $f$ is the operating frequency. Under Dynamic Voltage and Frequency Scaling (DVFS) \cite{mao2017survey}, the voltage scales approximately linearly with frequency, i.e., $V \propto f$. Therefore, the computation power consumption can be expressed as:
\begin{align}
P^{compute} = \kappa \cdot (f_{k_h})^3,
\end{align}
where $\kappa$ is the effective capacitance coefficient reflecting the processor technology characteristics. For a task executing $w_l$ CPU cycles, the required time is $\frac{w_l}{f_{k_h}}$, thus the total computation energy consumption is:
\begin{align}
E_{k_h,l}^{compute} = P^{compute} \cdot \frac{w_l}{f_{k_h}} = \kappa \cdot (f_{k_h})^2 \cdot w_l.
\end{align}

The result return energy consumption $E_{k,l}^{return}$ represents the energy consumption for transmitting computation results along the reverse path, also including both transmission and reception components:
\begin{align}
\scalebox{0.9}{$E_{k,l}^{return} = \sum_{i=h}^{2} \left( P_{k_i}^{tx} \cdot \frac{d_l}{R_{k_i,k_{i-1}}^{inter}} + P_{k_{i-1}}^{rx} \cdot \frac{d_l}{R_{k_i,k_{i-1}}^{inter}} \right)$},
\end{align}
where $P_{k_i}^{tx}$ is the transmission power of sending UAV $u_{k_i}$ and $P_{k_{i-1}}^{rx}$ is the reception power of receiving UAV $u_{k_{i-1}}$.

The downlink communication energy consumption $E_{k,l}^{downlink}$ represents the energy consumption for the serving UAV to transmit final results to IoT devices:
\begin{align}
E_{k,l}^{downlink} = P_k^{tx} \cdot T_l^{downlink},
\end{align}
where $P_k^{tx}$ is the air-to-ground communication transmission power of the UAV.

\subsection{Problem Formulation}
Based on the aforementioned models, this section formulates the joint optimization problem for UAV-assisted mobile edge computing systems. The core objective of the system is to minimize the comprehensive cost that simultaneously considers task completion time and UAV energy consumption as two key performance indicators, while satisfying QoS requirements.

\subsubsection{Decision Variables}
The optimization problem involves the following main decision variables:

\textbf{UAV Velocity Decisions}: $\mathbf{V} = \{v_{x,k}(t), v_{y,k}(t) \mid k \in [1,K], t \in [1,T]\}$ represents the velocity decisions of all UAVs within the time window.

\textbf{Processing Path Decisions}: $\mathbf{P} = \{\mathcal{P}_l \mid l \in \mathcal{L}\}$ represents the complete processing path for each task, where $\mathcal{P}_l = \{u_{k_1}, u_{k_2}, ..., u_{k_h}\}$. The first UAV $u_{k_1}$ in the path is the serving UAV that receives the task, and the last UAV $u_{k_h}$ is the final UAV that executes the computation. When $|\mathcal{P}_l| = 1$, it represents local processing.

\subsubsection{Objective Function}
The optimization objective of the system is to minimize the weighted combination of task completion time and UAV energy consumption:
\begin{align}
\label{eq: obj}
\min_{\mathbf{V}, \mathbf{P}} \quad F_{total} = \alpha \cdot \bar{F}_{time} + \beta \cdot \bar{F}_{energy},
\end{align}
where $\alpha$ and $\beta$ represent the weights for time and energy. $\bar{F}_{time}$ and $\bar{F}_{energy}$ represent the normalized time and energy costs, respectively.

To effectively compare time and energy objectives with different dimensions, we adopt a min-max normalization method. The normalized objective functions are defined as:
\begin{align}
&\bar{F}_{time} = \frac{F_{time} - F_{time}^{min}}{F_{time}^{max} - F_{time}^{min}}, \\
&\bar{F}_{energy} = \frac{F_{energy} - F_{energy}^{min}}{F_{energy}^{max} - F_{energy}^{min}},
\end{align}
where the time cost function is:
\begin{align}
F_{time} = \frac{1}{|\mathcal{L}|} \sum_{l \in \mathcal{L}} T_l^{total},
\end{align}
representing the average task completion time across all completed tasks in an episode.

The energy cost function is:
\begin{align}
F_{energy} = \frac{1}{K} \sum_{k=1}^{K} E_k^{total},
\end{align}
representing the average energy consumption per UAV in an episode.

$F_{time}^{min}$ and $F_{time}^{max}$ are the minimum and maximum average task completion times observed across all episodes and algorithms during training. Similarly, $F_{energy}^{min}$ and $F_{energy}^{max}$ are the minimum and maximum per-UAV energy consumption values. This normalization ensures both objectives are scaled to the range $[0, 1]$, enabling fair weighted combination in the optimization objective.

\subsubsection{Constraints}
\textbf{UAV Velocity Constraints}: UAV speed cannot exceed the maximum velocity:
\begin{align}
\label{ct: tra1}
||v_k(t)||_2 \leq v_k^{max}, \quad \forall k, t.
\end{align}

\noindent \textbf{UAV Position Update}: UAV position follows kinematic relationship:
\begin{align}
\label{ct: tra2}
pos_k(t+1) = pos_k(t) + v_k(t) \cdot \Delta t, \quad \forall k, t,
\end{align}
where $\Delta t$ is the time step duration. 

\noindent \textbf{Time Discretization Constraint}: The time step $\Delta t$ must ensure realistic UAV acceleration limits:
\begin{align}
\label{ct: time}
\Delta t \geq \max_{k \in \mathcal{U}} \frac{2 v_k^{max}}{a_k},
\end{align}
where $a_k$ is the acceleration capability of UAV $u_k$.

\noindent \textbf{Processing Path Constraints}: Each task must have one and only one processing path:
\begin{align}
\label{ct: assign}
|\mathcal{P}_l| \geq 1, \quad \forall l \in \mathcal{L}.
\end{align}

\noindent \textbf{Service Admission Constraints}: The first 
UAV in the path must be within communication range of the 
source device:
\begin{align}
\label{ct: comm}
I_{k_1,m}^{cov}(t) = 1, \quad \forall l \in \mathcal{L}_m, 
\mathcal{P}_l = \{u_{k_1}, ...\},
\end{align}
where $\mathcal{L}_m$ represents the task set generated 
by device $d_m$.

\noindent \textbf{UAV Energy Constraints}: The total energy consumption cannot exceed the available battery energy:
\begin{align}
\label{ct: eng}
E_k^{total} \leq energy_k(0), \quad \forall k \in \mathcal{U}
\end{align}
where $energy_k(0)$ represents the initial battery capacity of UAV $u_k$.

\noindent \textbf{UAV Computational Capacity Constraints}: The computational load must not exceed UAV capacity:
\begin{align}
\label{ct: comp}
load_k(t) \leq load_k^{max}, \quad \forall k \in \mathcal{U}, t
\end{align}
where $load_k^{max}$ represents the maximum computational capacity of UAV $u_k$.

\noindent \textbf{Path Connectivity Constraints}: Adjacent UAVs in the task processing path must have effective communication links:
\begin{align}
\label{ct: path}
d_{k_i,k_{i+1}} \leq R_{comm}, \quad \forall u_{k_i}, u_{k_{i+1}} \in \mathcal{P}_l,
\end{align}
where $R_{comm}$ is the maximum effective distance for inter-UAV communication.

\noindent \textbf{Deadline Constraints}: Each task should be completed before its deadline:
\begin{align}
\label{ct: ddl}
T_l^{total} \leq c_l, \quad \forall l \in \mathcal{L}.
\end{align}

\noindent \textbf{Coverage Constraints}: Each IoT device 
should be covered by at least one UAV to ensure service 
availability:
\begin{align}
\label{ct: cov}
\sum_{k \in \mathcal{U}} I_{k,m}^{cov}(t) \geq 1, 
\quad \forall m \in \mathcal{D}, t.
\end{align}

This optimization problem is a complex Mixed-Integer Nonlinear Programming (MINLP) problem involving joint optimization of continuous variables (UAV velocities) and discrete variables (path selection). The complexity of the problem stems from the coupling of UAV mobility, dynamic task arrivals, multi-hop cooperative processing, and other factors, requiring the design of efficient approaches to obtain near-optimal solutions.

\section{Proposed AirFed Framework}\label{sec:airfed}
\begin{figure*}[t]
\centering
\includegraphics[width=1.0\textwidth]{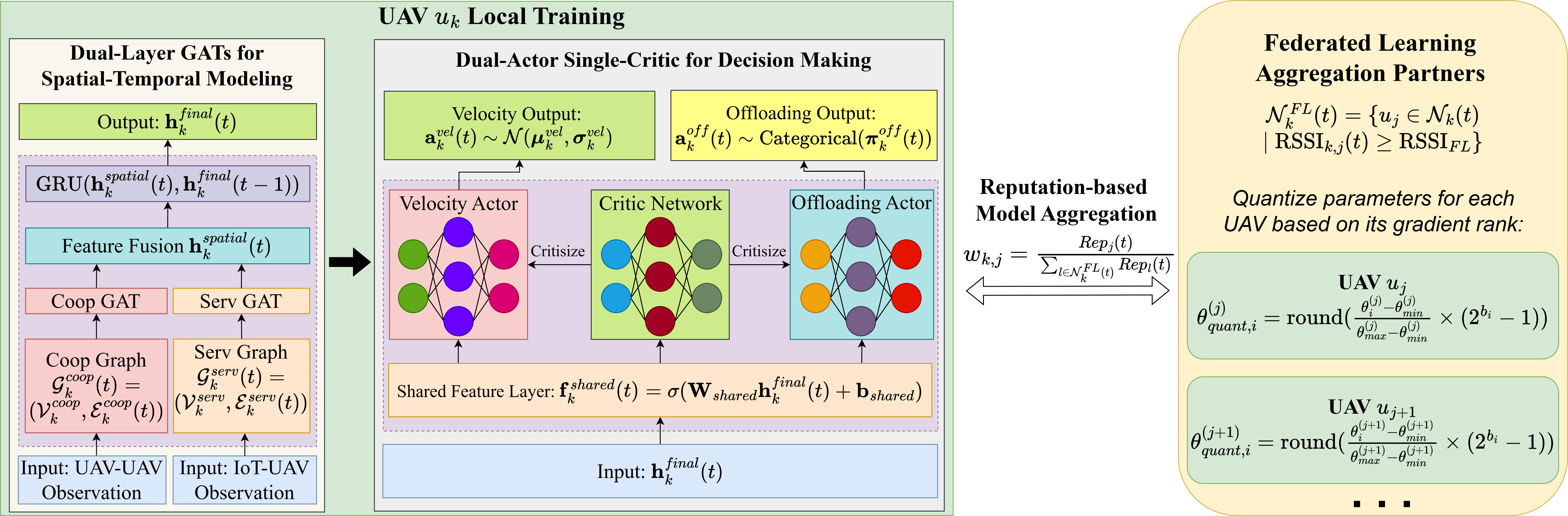}
\caption{Overall architecture of AirFed framework. The framework integrates dual-layer GATs for spatial-temporal modeling, dual-Actor single-Critic for hybrid decision making, and decentralized federated learning featuring reputation-based aggregation and gradient-sensitive quantization.}
\label{fig:airfed_architecture}
\end{figure*}

We now present AirFed, our proposed framework for multi-UAV cooperative edge computing. As illustrated in Fig.~\ref{fig:airfed_architecture}, AirFed consists of three key components: spatial-temporal feature extraction using dual-layer GATs, hybrid decision making via dual-actor single-critic architecture, and decentralized effiecient federated learning through reputation-based aggregation and adaptive quantization. Each UAV operates autonomously while collaborating with neighbors to improve collective performance. The following subsections detail each component.

\subsection{Dynamic Graph Attention Networks for Spatial-Temporal Modeling}
This section designs the dynamic GATs to model the complex spatial-temporal relationships in UAV-IoT systems. The architecture captures UAV cooperation relationships and UAV-IoT service relationships through a dual-layer graph structure, and achieves real-time network situational awareness through distributed graph attention computation.

\subsubsection{Dual-Layer Dynamic Graph Construction}
To comprehensively model the complex interaction relationships in UAV-assisted edge computing systems, we design a dual-layer dynamic graph structure. Each UAV $u_k$ maintains a local dual-layer graph $\mathcal{G}_k^{local}(t) = \{\mathcal{G}_k^{coop}(t), \mathcal{G}_k^{serv}(t)\}$, modeling cooperation relationships and service relationships respectively.

The UAV cooperation layer graph $\mathcal{G}_k^{coop}(t) = (\mathcal{V}_k^{coop}, \mathcal{E}_k^{coop}(t))$ models the communication and cooperation relationships between UAV $u_k$ and its neighbors. The node set $\mathcal{V}_k^{coop} = \{u_k\} \cup \mathcal{N}_k(t)$ contains the UAV itself and its communication neighbors, with the edge set defined as $(u_k, u_{k'}) \in \mathcal{E}_k^{coop}(t)$ if and only if $d_{k,k'}(t) \leq R_{comm}$. Each UAV node feature vector is:
\begin{align}
\scalebox{0.95}{$\mathbf{h}_{k'}^{coop}(t) = [pos_{k'}(t), v_{k'}(t), energy_{k'}(t), load_{k'}(t), f_{k'}]$},
\end{align}
containing position coordinates, flight velocity, remaining energy, current computational load, and computational capacity information. The cooperation edge feature vector is:
\begin{align}
\mathbf{e}_{k,k'}^{coop}(t) = [d_{k,k'}(t), RSSI_{k,k'}(t), B_{inter}, \eta_{k,k'}],
\end{align}
where $d_{k,k'}(t)$ represents the Euclidean distance between UAVs; $RSSI_{k,k'}(t)$ is the communication signal strength; $B_{inter}$ is the available bandwidth; $\eta_{k,k'}$ is the historical cooperation frequency.

The UAV-IoT service layer graph $\mathcal{G}_k^{serv}(t) = (\mathcal{V}_k^{serv}, \mathcal{E}_k^{serv}(t))$ models the coverage and service relationships between UAV $u_k$ and IoT devices. The node set $\mathcal{V}_k^{serv} = \{u_k\} \cup \mathcal{D}_k^{cov}(t)$ contains the UAV itself and its covered IoT devices. The UAV node uses the same feature representation as in the UAV cooperation layer, and the IoT device node features are:
\begin{align}
\mathbf{h}_m^{serv}(t) = [loc_m, |\mathcal{Q}_m(t)|, \lambda_m, \bar{c}_m(t)],
\end{align}
where $loc_m$ is the device location coordinates; $|\mathcal{Q}_m(t)|$ is the task queue length; $\lambda_m$ is the task generation rate; $\bar{c}_m(t)$ is the deadline of the most urgent task. Service edge features are:
\begin{align}
\mathbf{e}_{k,m}^{serv}(t) = [d_{k,m}(t), RSSI_{k,m}(t), R_{k,m}^{up}(t)],
\end{align}
where $d_{k,m}(t)$ is the distance between UAV and IoT device; $RSSI_{k,m}(t)$ is the signal strength; $R_{k,m}^{up}(t)$ is the uplink capacity.

Considering UAV mobility and task dynamics, each UAV's local graph structure requires real-time updates:
\begin{align}
\mathcal{G}_k^{local}(t+\Delta t) = \text{Update}(\mathcal{G}_k^{local}(t)).
\end{align}
To balance real-time performance and computational overhead, we adopt an adaptive update strategy based on mobility speed, where each UAV determines its local graph update interval:
\begin{align}
\Delta t_k^{update} = \frac{\Delta t_{base}}{1 + \alpha_{speed} \cdot ||v_k(t)||_2},
\end{align}
where $\alpha_{speed}$ is the speed sensitivity parameter, and UAVs with higher mobility speeds have higher graph structure update frequencies.

\subsubsection{Spatial-Temporal Graph Attention Network}
We design multi-scale spatial-temporal GATs where each UAV performs distributed graph attention computation based on its local graph. The dual-layer GATs adopt inter-layer interaction:
\begin{align}
\mathbf{H}_k^{coop,(l+1)}&(t) = \sigma\left(\mathbf{A}_{att}^{coop}(t) \mathbf{H}_k^{coop,(l)}(t) \mathbf{W}^{coop,(l)} \right. \nonumber \\
&\left. + \mathbf{M}_k^{serv \to coop}(t) \mathbf{H}_k^{serv,(l)}(t) \mathbf{W}^{cross,(l)}\right),
\end{align}

\begin{align}
\mathbf{H}_k^{serv,(l+1)}(t) = \sigma\left(\mathbf{A}_{att}^{serv}(t) \mathbf{H}_k^{serv,(l)}(t) \mathbf{W}^{serv,(l)}\right),
\end{align}
where $\mathbf{A}_{att}$ is the attention-based dynamic adjacency matrix, and the information transfer matrix $\mathbf{M}_k^{serv \to coop}(t)$ transfers task urgency information from the service layer to the cooperation layer:
\begin{align}
\mathbf{M}_{k,m}^{serv \to coop}(t) = \frac{I_{k,m}^{cov}(t) \cdot w_{k,m}(t)}{\sum_{m' \in \mathcal{D}_k^{cov}(t)} I_{k,m'}^{cov}(t) \cdot w_{k,m'}(t)},
\end{align}
where the weight function $w_{k,m}(t) = \exp(-\gamma_{urg} \cdot \bar{c}_m(t))$ is determined based on task urgency.

To adaptively learn the cooperation relationships between UAVs and the service relationships between UAVs and IoT devices, we design a multi-head attention mechanism. The cooperation layer attention weights are computed as:
\begin{align}
\scalebox{0.9}{${Attention}_{k,k'}^{coop,(h)}(t) = \frac{\exp(\mathbf{q}_k^{coop,(h)T} \mathbf{k}_{k'}^{coop,(h)} / \sqrt{d_k})}{\sum_{k'' \in \mathcal{N}_k(t)} \exp(\mathbf{q}_k^{coop,(h)T} \mathbf{k}_{k''}^{coop,(h)} / \sqrt{d_k})}$}.
\end{align}
The service layer adopts a similar attention computation. Through this design, UAVs can dynamically adjust their attention to different neighboring UAVs and IoT devices based on multi-dimensional information including geographical distance, load status, communication quality, and task urgency, achieving adaptive cooperation decisions.

To capture the temporal evolution patterns of the system, we model temporal dependencies through Gated Recurrent Unit (GRU) networks:
\begin{align}
\mathbf{h}_k^{final}(t) = \text{GRU}(\mathbf{h}_k^{spatial}(t), \mathbf{h}_k^{final}(t-1)),
\end{align}
where $\mathbf{h}_k^{spatial}(t)$ is the spatial feature representation fusing information from $\mathbf{H}_k^{coop}$ and $\mathbf{H}_k^{serv}$. The GRU is chosen for its computational efficiency while maintaining effective temporal modeling capabilities, which is crucial for real-time UAV edge computing applications. This dynamic GATs provides each UAV with a comprehensive state representation $\mathbf{h}_k^{final}(t)$ containing local status, neighborhood context, task urgency, and cooperation opportunities, serving as high-quality decision input for subsequent CMARL.

\subsection{Constrained Multi-Agent Reinforcement Learning}
This section reformulates the original optimization problem as a Constrained Multi-Agent Reinforcement Learning (CMARL) problem and designs a policy network architecture for handling hybrid action spaces. Each UAV acts as an independent agent, making joint decisions on trajectory planning and task offloading based on the state representation $\mathbf{h}_k^{final}(t)$ provided by GATs.

\subsubsection{Problem Reformulation as CMAMDP}
We reformulate the MINLP problem from Section \ref{system_model} as a Constrained Multi-Agent Markov Decision Process (CMAMDP) $\langle \mathcal{S}, \mathcal{A}, \mathcal{P}, \mathcal{R}, \mathcal{C} \rangle$. The system state space $\mathcal{S}$ contains comprehensive state representations of all UAVs provided by GATs:
\begin{align}
\mathcal{S} = \{\mathbf{h}_k^{final}(t) \mid k = 1, 2, ..., K\}.
\end{align}
Each UAV agent faces a hybrid action space $\mathcal{A}_k = \mathcal{A}_k^{vel} \times \mathcal{A}_k^{off}$, where continuous velocity actions $\mathbf{a}_k^{vel} \in \mathcal{A}_k^{vel} \subset \mathbb{R}^2$ represent UAV movement decisions executed at each time step, and discrete task actions $\mathbf{a}_k^{off} \in \mathcal{A}_k^{off}$ represent task offloading strategy executed upon task arrivals. The state transition probability $\mathcal{P}$ is determined by the joint actions executed by UAVs and environmental dynamics, and $\mathcal{R}$ is the reward function. According to physical feasibility and service requirements, the constraint set $\mathcal{C}$ is divided into hard constraints and soft constraints:
\begin{itemize}
    \item \textbf{Hard constraints} (must be strictly 
    satisfied): 
    \begin{align}
    \mathcal{C}_{hard} = \{\text{Eq. } \eqref{ct: tra1}, 
    \text{Eq. } \eqref{ct: tra2},  \text{Eq. } \eqref{ct: time}, \text{Eq. } \eqref{ct: assign}, \\
    \text{Eq. } \eqref{ct: comm}, \text{Eq. } \eqref{ct: eng}, 
    \text{Eq. } \eqref{ct: comp}, \text{Eq. } \eqref{ct: path}\},
    \end{align}
    strictly enforced during system operation.
    
    \item \textbf{Soft constraints} (violations allowed 
    but penalized):
    \begin{align}
    \mathcal{C}_{soft} = \{\text{Eq. } \eqref{ct: ddl}, 
    \text{Eq. } \eqref{ct: cov}\},
    \end{align}
    representing QoS requirements, handled through 
    penalty and reward terms in the reward function.
\end{itemize}

\subsubsection{Dual-Actor Single-Critic Architecture}
To simultaneously handle continuous trajectory optimization and discrete task offloading decisions, we design a dual-Actor single-Critic policy network architecture. This architecture unifies the processing of hybrid action spaces through shared feature extraction layers, ensuring coordination between the two types of decisions.

The shared feature layer converts GATs output state representations into decision features:
\begin{align}
\mathbf{f}_k^{shared}(t) = \sigma(\mathbf{W}_{shared} \mathbf{h}_k^{final}(t) + \mathbf{b}_{shared}).
\end{align}

The architecture employs two specialized Actor networks operating on the shared feature representation:
\begin{itemize}
    \item \textbf{Velocity Actor} outputs Gaussian distribution parameters for continuous velocity decisions, enabling smooth exploration-exploitation tradeoff through adjustable variance:
    \begin{align}
    \boldsymbol{\mu}_k^{vel}(t) &= \pi_{vel}(\mathbf{f}_k^{shared}(t); \theta_{vel}), \\
    \boldsymbol{\sigma}_k^{vel}(t) &= \sigma_{vel}(\mathbf{f}_k^{shared}(t); \theta_{vel}),
    \end{align}
    where velocity actions are sampled as $\mathbf{a}_k^{vel}(t) \sim \mathcal{N}(\boldsymbol{\mu}_k^{vel}, \boldsymbol{\sigma}_k^{vel})$ and constrained by velocity limits (Eq. \eqref{ct: tra1}) to ensure feasible UAV movement.
    
    \item \textbf{Offloading Actor} outputs probability distributions for discrete processing choices:
    \begin{align}
    \boldsymbol{\pi}_k^{off}(t) &= \text{softmax}(\pi_{off}(\mathbf{f}_k^{shared}(t); \theta_{off})), \\
    \mathbf{a}_k^{off}(t) &\sim \text{Categorical}(\boldsymbol{\pi}_k^{off}(t)),
    \end{align}
    allowing flexible selection between local computation and forwarding to neighboring UAVs based on current system states. Task offloading actions are executed only when new computational tasks arrive.
\end{itemize}

Both Actors share a unified Critic network for value estimation:
\begin{align}
V_k(t) = V_{critic}(\mathbf{f}_k^{shared}(t); \theta_{critic}),
\end{align}
providing a common baseline for policy gradient updates across both decision types.

\subsubsection{QoS-Aware Reward Design}
We design a multi-dimensional reward function that 
integrates the optimization objective with QoS guarantees:
\begin{align}
\scalebox{0.9}{$R_k(t) = R_k^{performance}(t) + R_k^{QoS}(t)$},
\end{align}
where $R_k^{QoS}(t) = R_k^{deadline}(t) + R_k^{coverage}(t)$ 
encompasses two key QoS aspects.

The performance reward corresponds to the negative 
incremental contribution to the optimization objective 
(Eq. \eqref{eq: obj}):
\begin{align}\label{eq:per}
R_k^{performance}(t) = -\Delta F_{total,k}(t).
\end{align}

The deadline reward imposes penalties on tasks that 
violate timing constraints:
\begin{align}\label{eq:ddl_r}
R_k^{deadline}(t) = -\lambda \max(0, T_l^{total} - c_l),
\end{align}
where $\lambda$ is the penalty weight for deadline 
violations.

The coverage reward incentivizes UAVs to maintain 
service availability for IoT devices:
\begin{align}\label{eq:cov_r}
R_k^{coverage}(t) = \eta \sum_{m \in \mathcal{D}} I_{k,m}^{cov}(t),
\end{align}
where $\eta$ is the coverage reward weight, ensuring 
adequate service coverage across the deployment area.

\subsubsection{Distributed Policy Optimization}
Each UAV independently maintains a complete dual-Actor single-Critic architecture and conducts distributed training. Both Actor networks are updated using policy gradient methods:
\begin{align}
&\nabla_{\theta_{vel}} J_{vel} = \mathbb{E}[\nabla_{\theta_{vel}} \log \pi_{\theta_{vel}}(\mathbf{a}_k^{vel}|\mathbf{s}_k) A_k(t)],
\end{align}
\begin{align}
&\nabla_{\theta_{off}} J_{off} = \mathbb{E}[\nabla_{\theta_{off}} \log \pi_{\theta_{off}}(\mathbf{a}_k^{off}|\mathbf{s}_k) A_k(t)],
\end{align}
where the advantage function $A_k(t) = R_k(t) + \gamma V_k(t+1) - V_k(t)$ provides unified training signals for both Actors.

The Critic network is updated through temporal difference error:
\begin{align}
L_{critic} = \mathbb{E}[(R_k(t) + \gamma V_k(t+1) - V_k(t))^2].
\end{align}

The total loss function integrates policy losses from both Actors and the value loss from the Critic:
\begin{align}
L_{total} = L_{vel} + L_{off} + L_{critic} + \beta_{entropy}H(\pi),
\end{align}
where $H(\pi)$ is the entropy regularization term that promotes sufficient policy exploration.

Algorithm \ref{alg:cmarl} summarizes the overall CMARL training process for UAV cooperative edge computing. The algorithm operates in four main phases: Lines 1-5 perform system initialization including network parameters and local graph structures. During online execution (Lines 9-27), each UAV adaptively updates its local dual-layer graph based on mobility speed (Lines 10-13), processes spatial-temporal features through GATs and GRU (Lines 15-17), and generates hybrid actions via the dual-actor architecture (Lines 20-27), with velocity decisions made continuously and task offloading triggered by task arrivals. The environment interaction phase (Lines 29-30) executes joint UAV velocity actions and task offloading decisions when applicable. The experience collection (Lines 31-38) computes multi-dimensional rewards including performance, constraint, and coverage components. Finally, distributed policy updates (Lines 40-55) enable each UAV to independently update its dual-actor and critic networks using policy gradients and temporal difference learning.

\begin{algorithm}[htbp]
\footnotesize
\caption{Constrained Multi-Agent Reinforcement Learning for UAV Cooperative Edge Computing}
\label{alg:cmarl}
\begin{algorithmic}[1]
\REQUIRE UAV set $\mathcal{U}$, IoT device set $\mathcal{D}$, communication range $R_{comm}$, Network parameters $\theta_{vel}$, $\theta_{off}$, $\theta_{critic}$, learning rates $\alpha_{vel}$, $\alpha_{off}$, $\alpha_{critic}$

\STATE // \textbf{Initialization}
\STATE Initialize GATs parameters and network weights
\STATE Initialize UAV positions $pos_k(0)$ and states $s_k(0)$ for all $k \in \mathcal{U}$
\STATE Initialize experience buffer $\mathcal{B}_k$ for each UAV $k$
\STATE Initialize local graphs $\mathcal{G}_k^{local}(0)$ for all UAVs

\FOR{training iteration $e = 1$ to $E$}
    \FOR{time step $t = 1$ to $T$}
        \FOR{each UAV $k \in \mathcal{U}$}
            \STATE // \textbf{Adaptive Graph Update}
            \STATE Compute update interval: $\Delta t_k^{update} = \frac{\Delta t_{base}}{1 + \alpha_{speed} \cdot ||v_k(t)||_2}$
            \IF{$t \mod \Delta t_k^{update} = 0$}
                \STATE Update local dual-layer graph $\mathcal{G}_k^{local}(t) = \{\mathcal{G}_k^{coop}(t), \mathcal{G}_k^{serv}(t)\}$
            \ENDIF
            
            \STATE // \textbf{Feature Processing}
            \STATE Apply GATs on $\mathcal{G}_k^{local}(t)$ to obtain $\mathbf{h}_k^{spatial}(t)$
            \STATE Update temporal state: $\mathbf{h}_k^{final}(t) = \text{GRU}(\mathbf{h}_k^{spatial}(t), \mathbf{h}_k^{final}(t-1))$
            \STATE Compute shared decision features: $\mathbf{f}_k^{shared}(t) = \sigma(\mathbf{W}_{shared} \mathbf{h}_k^{final}(t) + \mathbf{b}_{shared})$
            
            \STATE // \textbf{Dual-Actor Action Selection}
            \STATE // Velocity Actor
            \STATE $\boldsymbol{\mu}_k^{vel}(t) = \pi_{vel}(\mathbf{f}_k^{shared}(t); \theta_{vel})$
            \STATE $\boldsymbol{\sigma}_k^{vel}(t) = \sigma_{vel}(\mathbf{f}_k^{shared}(t); \theta_{vel})$
            \STATE Sample $\mathbf{a}_k^{vel}(t) \sim \mathcal{N}(\boldsymbol{\mu}_k^{vel}, \boldsymbol{\sigma}_k^{vel})$
            
            \STATE // Offloading Actor  
            \STATE $\boldsymbol{\pi}_k^{off}(t) = \text{softmax}(\pi_{off}(\mathbf{f}_k^{shared}(t); \theta_{off}))$
            \STATE Sample $\mathbf{a}_k^{off}(t) \sim \text{Categorical}(\boldsymbol{\pi}_k^{off}(t))$
            
            \STATE // Critic Value Estimation
            \STATE $V_k(t) = V_{critic}(\mathbf{f}_k^{shared}(t); \theta_{critic})$
        \ENDFOR
        
        \STATE // \textbf{Environment Interaction}
        \STATE Execute joint actions $\{\mathbf{a}_k^{vel}(t), \mathbf{a}_k^{off}(t)\}_{k \in \mathcal{U}}$
        
        \STATE // \textbf{Experience Collection}
        \FOR{each UAV $k \in \mathcal{U}$}
            \STATE Compute performance reward: $R_k^{performance}(t) = -\Delta F_{total,k}(t)$
            \STATE Compute constraint penalty: $R_k^{constraint}(t) = -\lambda \max(0, T_l^{total} - c_l)$
            \STATE Compute coverage reward: $R_k^{coverage}(t) = \eta \sum_{m \in \mathcal{D}} I_{k,m}^{cov}(t)$
            \STATE Total reward: $R_k(t) = R_k^{performance}(t) + R_k^{constraint}(t) + R_k^{coverage}(t)$
            \STATE Store transition $(\mathbf{h}_k^{final}(t), \mathbf{a}_k^{vel}(t), \mathbf{a}_k^{off}(t), R_k(t), \mathbf{h}_k^{final}(t+1))$ in $\mathcal{B}_k$
        \ENDFOR
    \ENDFOR
    
    \STATE // \textbf{Distributed Policy Updates}
    \FOR{each UAV $k \in \mathcal{U}$}
        \STATE Sample mini-batch from experience buffer $\mathcal{B}_k$
        
        \STATE // Advantage Computation
        \FOR{each transition in mini-batch}
            \STATE Compute advantage: $A_k(t) = R_k(t) + \gamma V_k(t+1) - V_k(t)$
        \ENDFOR
        
        \STATE // Actor Networks Update
        \STATE Compute velocity policy gradient: $\nabla_{\theta_{vel}} J_{vel} = \mathbb{E}[\nabla_{\theta_{vel}} \log \pi_{\theta_{vel}}(\mathbf{a}_k^{vel}|\mathbf{s}_k) A_k(t)]$
        \STATE Update: $\theta_{vel} \leftarrow \theta_{vel} + \alpha_{vel} \nabla_{\theta_{vel}} J_{vel}$
        
        \STATE Compute task policy gradient: $\nabla_{\theta_{off}} J_{off} = \mathbb{E}[\nabla_{\theta_{off}} \log \pi_{\theta_{off}}(\mathbf{a}_k^{off}|\mathbf{s}_k) A_k(t)]$
        \STATE Update: $\theta_{off} \leftarrow \theta_{off} + \alpha_{off} \nabla_{\theta_{off}} J_{off}$
        
        \STATE // Critic Network Update
        \STATE Compute value loss: $L_{critic} = \mathbb{E}[(R_k(t) + \gamma V_k(t+1) - V_k(t))^2]$
        \STATE Update: $\theta_{critic} \leftarrow \theta_{critic} - \alpha_{critic} \nabla_{\theta_{critic}} L_{critic}$
    \ENDFOR
\ENDFOR

\RETURN Trained policy networks $\pi_{vel}$, $\pi_{off}$, and value network $V_{critic}$
\end{algorithmic}
\end{algorithm}

\subsection{Multi-UAV Decentralized Federated Learning}
While the CMARL enables each UAV to learn policies independently, the lack of knowledge sharing mechanisms among UAVs limits the overall learning efficiency of the system. We adopt a decentralized federated learning approach where each UAV serves both as a DRL agent and a federated learning participant. The federated learning parameter set for UAV $u_k$ is:
\begin{align}
\Theta_k^{FL} = \{\theta_{vel}^{(k)}, \theta_{off}^{(k)}, \theta_{critic}^{(k)}\},
\end{align}
where $\theta_{vel}^{(k)}$ represents the velocity decision network parameters, $\theta_{off}^{(k)}$ denotes the task offloading decision network parameters, $\theta_{critic}^{(k)}$ indicates the value estimation network parameters. The federated learning process leverages the UAV cooperation layer graph $\mathcal{G}_k^{coop}(t)$ to establish communication topology for distributed parameter exchange, enabling collaborative learning without centralized coordination.

\subsubsection{Reputation-based Model Aggregation}
UAV $u_k$ selects aggregation partners from its communication neighbors:
\begin{align}
\mathcal{N}_k^{FL}(t) = \{u_j \in \mathcal{N}_k(t) \mid \text{RSSI}_{k,j}(t) \geq \text{RSSI}_{FL}\},
\end{align}
where $\text{RSSI}_{FL}$ is the minimum signal strength threshold required for federated learning.

To evaluate the trustworthiness and contribution value of different UAVs, we design a reputation mechanism based on historical performance. Considering the characteristics of UAV systems, reputation assessment needs to reflect two key aspects: the reliability of UAVs in executing computational tasks and the stability of UAVs in participating in federated learning. The former ensures learning from high-quality data and experiences, while the latter guarantees the continuity and effectiveness of parameter exchange.

The task execution reputation of UAV $j$ is defined as:
\begin{align}
Succ_j(t) = \frac{\text{successfully completed tasks}}{\text{total assigned tasks}},
\end{align}
where the numerator represents the number of tasks successfully completed by UAV $j$ within deadlines, and the denominator represents the total number of tasks assigned to UAV $j$. This metric reflects the computational capability and task processing reliability of the UAV.

The communication reputation of UAV $j$ is defined as:
\begin{align}
Stab_j(t) = \frac{\text{successful FL communications}}{\text{total FL attempts}},
\end{align}
where the numerator represents the number of successful federated learning parameter exchanges completed by UAV $j$, and the denominator represents the total number of federated learning communication attempts. This metric reflects the network stability and collaborative reliability of the UAV.

At each time step, we first calculate the instantaneous reputation based on current performance metrics:
\begin{align}
\tilde{Rep}_j(t) = \alpha_{succ} \cdot Succ_j(t) + \alpha_{stab} \cdot Stab_j(t),
\end{align}
where $\alpha_{succ} + \alpha_{stab} = 1$ are weight parameters balancing task execution and communication reliability.

To prevent abrupt reputation changes caused by temporary performance fluctuations, we apply exponential moving average to smooth the reputation updates:
\begin{align}
Rep_j(t) = \rho \cdot Rep_j(t-1) + (1-\rho) \cdot \tilde{Rep}_j(t),
\end{align}
where $\rho \in [0,1]$ is the forgetting factor controlling the influence of historical reputation versus current performance. And the reputation-based aggregation weights are defined as:
\begin{align}
w_{k,j} = \frac{Rep_j(t)}{\sum_{l \in \mathcal{N}_k^{FL}(t)} Rep_l(t)}.
\end{align}

\subsubsection{Communication-Efficient Adaptive Updates}
Due to the typically much smaller bandwidth of inter-UAV communication compared to terrestrial networks, directly transmitting full-precision model parameters would incur enormous communication overhead. To reduce communication burden while maintaining learning effectiveness, we propose a gradient-sensitive adaptive quantization mechanism. According to the first-order Taylor expansion of the loss function, the gradient magnitude $|\partial \mathcal{L} / \partial \theta_i|$ directly reflects the impact of quantization error in parameter value on the loss function \cite{lee2021network}. Parameters with large gradient magnitudes are more sensitive to quantization errors and require higher transmission precision, while parameters with small gradient magnitudes can tolerate more aggressive quantization compression \cite{lee2021network}. Based on this observation, we apply high-bit quantization to parameters with large gradient magnitudes and low-bit quantization to parameters with small gradient magnitudes.

Specifically, when UAV $u_j$ prepares to transmit parameters, it computes the gradient rank $r_i = \text{rank}(g_i)$ for each parameter based on the gradient absolute value $g_i = |\partial \mathcal{L} / \partial \theta_i|$, where the rank function assigns values in descending order. Based on the gradient rank, the quantization bit width for each parameter is determined as:
\begin{align}
b_i = b_{min} + \left\lfloor \left(1 - \frac{r_i - 1}{n_p - 1}\right) \times (b_{max} - b_{min}) \right\rfloor,
\end{align}
where $b_{min}$ and $b_{max}$ are the minimum and maximum bit widths respectively, defining the range of quantization precision, and $n_p$ is the total number of network parameters. This mapping ensures that parameters with higher gradient ranks receive high-precision quantization close to $b_{max}$, while parameters with lower gradient ranks receive low-precision quantization close to $b_{min}$. The parameter quantization process is:
\begin{align}
\theta_{quant,i} = \text{round}\left(\frac{\theta_i - \theta_{min}}{\theta_{max} - \theta_{min}} \times (2^{b_i} - 1)\right),
\end{align}
where $\theta_{max}$ and $\theta_{min}$ are the extreme values of the parameter vector. During transmission, the bit width identifiers $\{b_i\}$ of each parameter, along with $\theta_{max}$ and $\theta_{min}$, are attached to the quantized parameters. Upon reception, the receiving UAV reconstructs the full-precision parameters through dequantization:
\begin{align}
\theta_i = \theta_{min} + \frac{\theta_{quant,i}}{2^{b_i} - 1} \times (\theta_{max} - \theta_{min}),
\end{align}
which linearly maps the quantized integer values back to the original floating-point space. This mechanism allocates transmission precision according to the relative sensitivity of parameters, achieving a balance between parameter transmission efficiency and model convergence quality in resource-constrained UAV networks.

Considering the high mobility and dynamic network topology of UAVs, strict synchronous federated learning is difficult to achieve. We design an asynchronous update mechanism that allows different UAVs to perform model aggregation at different frequencies. The aggregation frequency of each UAV $k$ is adaptively adjusted according to its movement speed:
\begin{align}
f_{agg}^{(k)} = f_{base} \times (1 + \alpha_{mobility} \times ||v_k(t)||_2),
\end{align}
where $f_{base}$ is the baseline aggregation frequency, and $\alpha_{mobility}$ is the mobility sensitivity parameter. The rationale behind this design is that fast-moving UAVs experience more frequent changes in network neighbors and require more timely model updates to adapt to new collaborative environments, while relatively stationary UAVs can adopt lower aggregation frequencies to reduce communication overhead.

When UAVs move beyond communication range or encounter signal interference, the federated learning process needs to handle communication failures. We adopt a best-effort aggregation strategy: UAV $k$ collects neighbor parameters within a predetermined time window and performs aggregation based on received parameters after timeout. Each UAV includes its own locally trained parameters in the aggregation process, with all participating UAVs (including itself) weighted by their respective reputations. Aggregation weights are normalized according to actually participating neighbors and the UAV itself:
\begin{align}
w_{k,j}^{normalized} = \frac{Rep_j(t)}{\sum_{l \in \{k\} \cup \mathcal{N}_k^{received}(t)} Rep_l(t)},
\end{align}
where $\mathcal{N}_k^{received}(t) \subseteq \mathcal{N}_k^{FL}(t)$ is the set of neighbors from which parameters were successfully received. This fault-tolerance mechanism ensures that the federated learning process can continue under partial communication failures while preserving local training outcomes, improving system robustness. The parameter aggregation update rules are:
\begin{align}
\theta_{vel}^{(k)} &\leftarrow \sum_{j \in \{k\} \cup \mathcal{N}_k^{received}(t)} w_{k,j}^{normalized} \theta_{vel}^{(j)}, \\
\theta_{off}^{(k)} &\leftarrow \sum_{j \in \{k\} \cup \mathcal{N}_k^{received}(t)} w_{k,j}^{normalized} \theta_{off}^{(j)}, \\
\theta_{critic}^{(k)} &\leftarrow \sum_{j \in \{k\} \cup \mathcal{N}_k^{received}(t)} w_{k,j}^{normalized} \theta_{critic}^{(j)}.
\end{align}

Algorithm \ref{alg:federated_learning} presents the decentralized federated learning for UAV cooperative edge computing. The algorithm operates through three concurrent processes that run independently for each UAV. The local CMARL training process (Lines 4-6) continuously executes DRL and updates local parameters. The reputation management (Lines 7-10) operates in an event-driven manner, updating task execution and communication reliability statistics, computing instantaneous reputation values, and applying exponential moving average smoothing to maintain stable reputation assessments. The core federated learning process (Lines 11-30) implements adaptive aggregation scheduling based on UAV mobility, where fast-moving UAVs aggregate more frequently to adapt to changing network topologies (Line 12). During each aggregation cycle, UAVs select qualified neighbors based on signal strength thresholds (Line 14), apply gradient-sensitive adaptive quantization to compress parameters based on gradient sensitivity (Lines 16-20), and send quantized parameters along with bit width identifiers and normalization bounds (Line 21). Upon reception, UAVs reconstruct full-precision parameters through dequantization (Lines 23-25) and perform reputation-weighted aggregation that includes both locally trained parameters and received neighbor parameters, with weights normalized across all participants (Lines 26-29). This decentralized approach enables knowledge sharing among UAVs while maintaining communication efficiency and robustness against communication failures and dynamic network conditions.

\begin{algorithm}[t]
\footnotesize
\caption{Decentralized Federated Learning for UAV Cooperative Edge Computing}
\label{alg:federated_learning}
\begin{algorithmic}[1]
\REQUIRE UAV set $\mathcal{U}$, reputation parameters $\alpha_{succ}$, $\alpha_{stab}$, $\rho$, FL parameters $RSSI_{FL}$, $b_{min}$, $b_{max}$, $f_{base}$, $\alpha_{mobility}$, $T_{timeout}$
\STATE // \textbf{Note: CMARL training, reputation updates, and FL aggregation run concurrently and independently}
\FOR{each UAV $k \in \mathcal{U}$ in parallel}
    \WHILE{system operating}
        \STATE // \textbf{Local CMARL Training (Continuous)}
        \STATE Execute local CMARL training (Algorithm \ref{alg:cmarl})
        \STATE Update local parameters $\{\boldsymbol{\theta}_{vel}^{(k)}, \boldsymbol{\theta}_{off}^{(k)}, \boldsymbol{\theta}_{critic}^{(k)}\}$
        
        \STATE // \textbf{Reputation Management (Event-driven)}
        \STATE Update performance statistics $Succ_k(t)$ and $Stab_k(t)$ based on task and communication events
        \STATE Compute instantaneous reputation: $\tilde{Rep}_k(t) = \alpha_{succ} \cdot Succ_k(t) + \alpha_{stab} \cdot Stab_k(t)$
        \STATE Update smoothed reputation: $Rep_k(t) = \rho \cdot Rep_k(t-1) + (1-\rho) \cdot \tilde{Rep}_k(t)$
        
        \STATE // \textbf{Adaptive FL Aggregation (Periodic)}
        \STATE Compute aggregation frequency: $f_{agg}^{(k)} = f_{base} \times (1 + \alpha_{mobility} \times ||v_k(t)||_2)$
        
        \IF{aggregation condition triggered based on $f_{agg}^{(k)}$}
            \STATE Select FL neighbors: $\mathcal{N}_k^{FL}(t) = \{u_j \in \mathcal{N}_k(t) \mid RSSI_{k,j}(t) \geq RSSI_{FL}\}$
            
            \STATE // \textbf{Parameter Exchange}
            \FOR{each neighbor $j \in \mathcal{N}_k^{FL}(t)$}
                \STATE UAV $j$ computes gradient rank: $r_i = \text{rank}(|\partial \mathcal{L} / \partial \theta_i^{(j)}|)$ in descending order
                \STATE Determine bit width: $b_i = b_{min} + \lfloor (1 - \frac{r_i - 1}{n - 1}) \times (b_{max} - b_{min}) \rfloor$
                \STATE Quantize parameters: $\theta_{quant,i}^{(j)} = \text{round}(\frac{\theta_i^{(j)} - \theta_{min}^{(j)}}{\theta_{max}^{(j)} - \theta_{min}^{(j)}} \times (2^{b_i} - 1))$
                \STATE UAV $j$ sends $\{\boldsymbol{\theta}_{quant}^{(j)}, \{b_i\}, \theta_{max}^{(j)}, \theta_{min}^{(j)}, Rep_j(t)\}$ to UAV $k$
            \ENDFOR
            
            \STATE // \textbf{Parameter Reconstruction and Aggregation}
            \FOR{each successfully received neighbor $j \in \mathcal{N}_k^{received}$}
                \STATE Dequantize: $\theta_i^{(j)} = \theta_{min}^{(j)} + \frac{\theta_{quant,i}^{(j)}}{2^{b_i} - 1} \times (\theta_{max}^{(j)} - \theta_{min}^{(j)})$
            \ENDFOR
            
            \STATE Compute normalized weights: $w_{k,j}^{normalized} = \frac{Rep_j(t)}{\sum_{l \in \{k\} \cup \mathcal{N}_k^{received}} Rep_l(t)}$ for $j \in \{k\} \cup \mathcal{N}_k^{received}$
            \STATE $\theta_{vel}^{(k)} \leftarrow \sum_{j \in \{k\} \cup \mathcal{N}_k^{received}} w_{k,j}^{normalized} \theta_{vel}^{(j)}$
            \STATE $\theta_{off}^{(k)} \leftarrow \sum_{j \in \{k\} \cup \mathcal{N}_k^{received}} w_{k,j}^{normalized} \theta_{off}^{(j)}$
            \STATE $\theta_{critic}^{(k)} \leftarrow \sum_{j \in \{k\} \cup \mathcal{N}_k^{received}} w_{k,j}^{normalized} \theta_{critic}^{(j)}$
        \ENDIF
    \ENDWHILE
\ENDFOR
\end{algorithmic}
\end{algorithm}

\section{Performance Evaluation}
\label{evaluation}
We evaluate AirFed through extensive experiments, analyzing convergence behavior, QoS guarantees, component contributions, and scalability across different system configurations.

\subsection{Experimental Setup}
We describe the evaluation environment, baseline approaches, and hyperparameter configurations.

\subsubsection{Simulation Environment and Parameters}
We implement AirFed using ReinFog~\cite{wang2025reinfog}, a modular DRL framework with native support for distributed heterogeneous agent deployment and flexible communication mechanisms, making it well-suited for our decentralized multi-UAV system. We leverage ReinFog's distributed learner infrastructure to deploy each UAV as an autonomous agent, comprising the GATs, trajectory/offloading actors, and the unified critic. Each UAV interacts with the environment, performs policy updates, and periodically exchanges model parameters with neighboring UAVs. To realize the federated learning protocol, we extend ReinFog's parameter synchronization mechanism by integrating reputation-weighted aggregation and gradient-sensitive adaptive quantization. Additionally, we develop a discrete-event simulator in Python following the system model in Section~\ref{system_model}, which models UAV mobility, channel dynamics, task arrivals, and resource constraints, working in conjunction with ReinFog to provide environmental feedback for training. Parameter configurations adhere to typical UAV-MEC deployment scenarios.

\textbf{System Deployment.} 
The service area spans $1000 \times 1000$ m$^2$ with $K=6$ UAVs operating at altitudes uniformly distributed in [80, 150] m \cite{shabanighazikelayeh2022optimal}. UAV initial positions are determined via k-means clustering on IoT device locations \cite{liu2020resource}. $M=40$ IoT devices are randomly distributed on the ground, simulating smart city sensor deployments \cite{samir2019uav}. For scalability analysis, we vary $K \in \{2,4,6,8,10\}$ and $M \in \{20,40,60,80,100\}$.

\textbf{UAV Specifications.} 
The computational frequency $f_k$ of each UAV is uniformly sampled from [1, 3] GHz, reflecting heterogeneous edge computing platforms \cite{valente2024heterogeneous}. Maximum flight velocity $v_k^{\max} = 20$ m/s and acceleration capability $a_k = 5$ m/s$^2$ follow \cite{ren2025safety, 7487284}. Initial battery capacity $energy_k(0) = 500$ kJ follows current lithium polymer battery specifications \cite{tyrovolas2022energy}. 

\textbf{Communication Environment.} 
Inter-UAV communication range $R_{\text{comm}} = 400$ m is based on 2.4 GHz WiFi air-to-air transmission characteristics \cite{10.1145/3345838.3356008}. RSSI threshold $\text{RSSI}_{\min} = -90$ dBm follows IEEE 802.11 receiver sensitivity standards \cite{10.1145/2789168.2790120}. Channel gains are $G_0 = -30$ dB (air-to-ground) and $G_{\text{inter}} = -20$ dB (air-to-air) following free-space path loss models \cite{9749282, 8108204}. Bandwidth allocation is $B = 10$ MHz for device-to-UAV links and $B_{\text{inter}} = 20$ MHz for inter-UAV links \cite{9126800, 9612717}. Thermal noise power is $N_0 = -114$ dBm, derived from standard noise spectral density and bandwidth \cite{6584940}.

\textbf{Power Configuration.} 
UAV transmission and reception power are $P_k^{\text{tx}} = 0.5$ W and $P_k^{\text{rx}} = 0.1$ W, with IoT device transmission power $P_m^{\text{tx}} = 0.1$ W \cite{yang2018energy, shang2019unmanned, tyrovolas2022energy}. Hovering power $P_{\text{hover}} = 80$ W is based on medium-sized quadrotor measurements carrying computational payloads \cite{liu2020path}. Aerodynamic parameters are air density $\rho = 1.225$ kg/m$^3$, effective drag area $A = 0.1 m^2$, and drag coefficient $C_D = 0.3$ \cite{dai2023energy, Wan2020}. Computational energy uses CMOS effective switched capacitance $\kappa = 10^{-18}$ \cite{ho2023limits}.

\textbf{Task Model.} 
IoT devices generate tasks following independent Poisson processes with rates $\lambda_m$ uniformly sampled from [0.3, 0.8] tasks/s \cite{chen2022optimal}. Task parameters include computational requirements $w_l \in [50, 200]$ Mcycles, input data $s_l \in [1, 3]$ MB, output data $d_l \in [0.1, 0.5]$ MB, and deadlines $c_l \in [5, 20]$ s, following \cite{cheng2019balanced, hu2020heterogeneous, 10141528, 10748391}.

\textbf{Simulation Protocol.} 
System performance is evaluated over 100 operational episodes, each spanning 300 s. Episode initialization employs k-means clustering on IoT device locations for UAV positioning. Each configuration undergoes 15 independent runs with different random seeds, with results reported as mean values.

\subsubsection{Baseline Approaches}
We compare our proposed AirFed against four state-of-the-art approaches:
\begin{itemize}
    \item \textbf{IF-DDPG} \cite{li2024computation}: A federated DDPG-based approach featuring dual experience replay and mixed noise exploration.
    
    \item \textbf{MUTO} \cite{ning2024multi}: A multi-agent Actor-Critic approach with prioritized experience replay for UAV trajectory optimization in differentiated service scenarios.
    
    \item \textbf{JTORA} \cite{chen2025joint}: A hybrid approach combining Lyapunov optimization for queue stability with SAC for UAV trajectory planning and resource allocation in dynamic MEC systems.
    
    \item \textbf{BCD-SCA} \cite{gao2025improving}: A iterative convex optimization combining BCD and SCA for joint trajectory and resource allocation.
\end{itemize}

These baselines collectively represent different solution paradigms: federated learning with independent agents (IF-DDPG), prioritized multi-agent DRL (MUTO), hybrid stochastic optimization with DRL (JTORA), and non-learning convex optimization (BCD-SCA). This selection enables comprehensive evaluation of AirFed across different methodological approaches. 

\renewcommand{\arraystretch}{0.8}
\begin{table}[h]
\centering
\footnotesize
\caption{Hyperparameter Configuration Summary}
\label{tab:hyperparams}
\begin{tabular}{lc}
\toprule
\textbf{Parameter} & \textbf{Value} \\
\midrule
\multicolumn{2}{l}{\textit{GATs Architecture}} \\
GATs layers & 2 \\
GATs hidden layers & [128, 64] \\
Attention heads & 4 \\
GRU hidden dimension & 128 \\
Speed sensitivity $\alpha_{speed}$ & 0.1 \\
Task urgency weight $\gamma_{urg}$ & 0.5 \\
\midrule
\multicolumn{2}{l}{\textit{Actor-Critic Architecture}} \\
Shared feature layer & 128 \\
Velocity Actor hidden layers & [128, 128] \\
Offloading Actor hidden layers & [128, 128] \\
Critic hidden layers & [128, 64] \\
Activation function & ReLU \\
\midrule
\multicolumn{2}{l}{\textit{DRL Training}} \\
Velocity Actor learning rate $\alpha_{vel}$ & $3 \times 10^{-4}$ \\
Offloading Actor learning rate $\alpha_{off}$ & $3 \times 10^{-4}$ \\
Critic learning rate $\alpha_{critic}$ & $5 \times 10^{-4}$ \\
Discount factor $\gamma$ & 0.95 \\
Optimizer & Adam \\
Entropy regularization $\beta_{entropy}$ & 0.01 \\
\midrule
\multicolumn{2}{l}{\textit{Reward Function Weights}} \\
Time weight $\alpha$ & 0.5 \\
Energy weight $\beta$ & 0.5 \\
Deadline penalty $\lambda$ & 10 \\
Coverage reward $\eta$ & 0.1 \\
\midrule
\multicolumn{2}{l}{\textit{Federated Learning}} \\
FL communication RSSI threshold & -85 dBm \\
Minimum bit width $b_{min}$ & 4 \\
Maximum bit width $b_{max}$ & 16 \\
Base aggregation frequency $f_{base}$ & 0.03 Hz \\
Mobility sensitivity $\alpha_{mobility}$ & 0.05 \\
Task success weight $\alpha_{succ}$ & 0.6 \\
Communication stability weight $\alpha_{stab}$ & 0.4 \\
Reputation smoothing factor $\rho$ & 0.75 \\
\bottomrule
\end{tabular}
\end{table}
\renewcommand{\arraystretch}{1}

\begin{figure*}[htbp]
\centering
\begin{subfigure}[b]{0.32\textwidth}
    \centering
    \includegraphics[width=\textwidth]{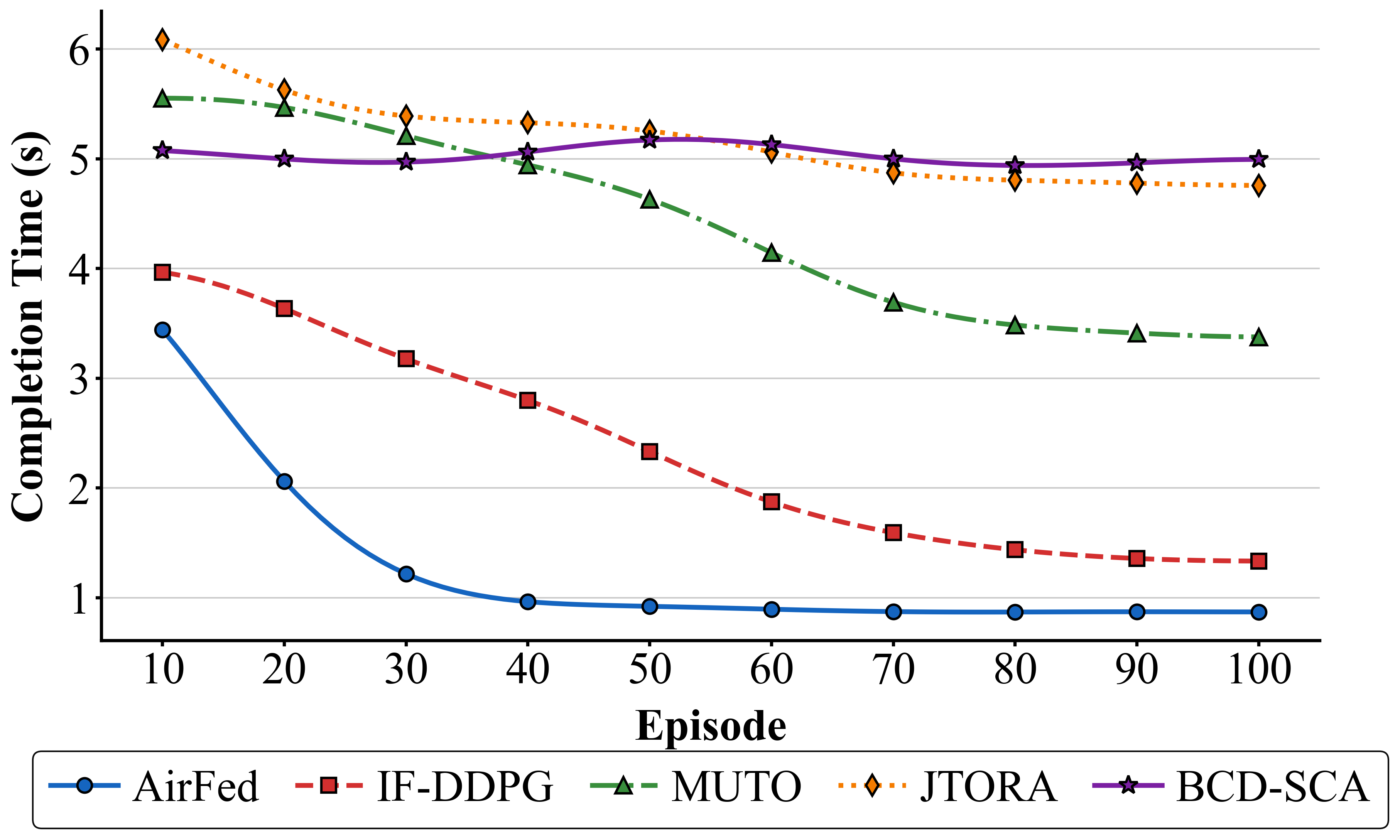}
    \caption{Task completion time}
    \label{fig:convergence_time}
\end{subfigure}
\hfill
\begin{subfigure}[b]{0.32\textwidth}
    \centering
    \includegraphics[width=\textwidth]{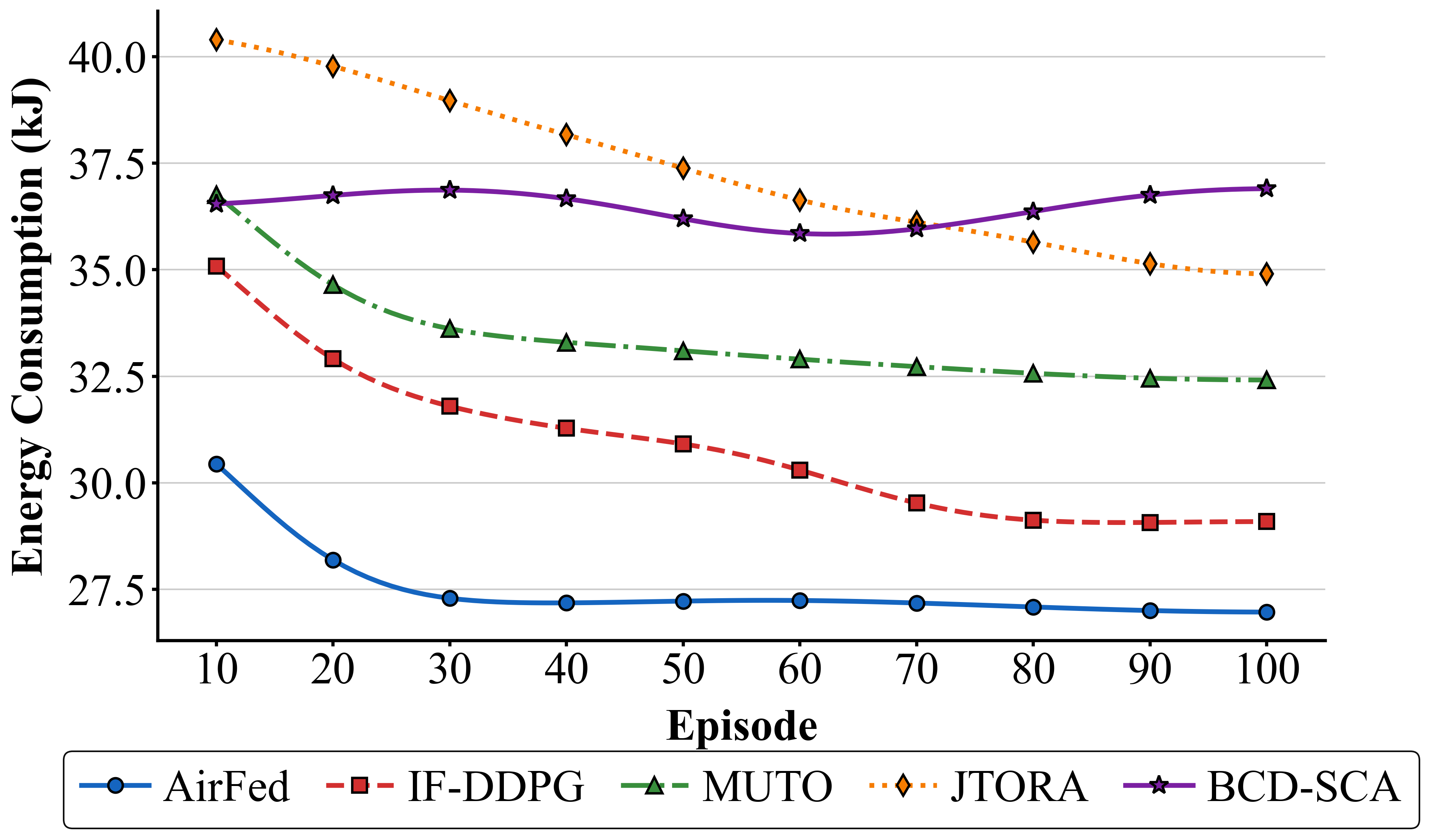}
    \caption{Energy consumption}
    \label{fig:convergence_energy}
\end{subfigure}
\hfill
\begin{subfigure}[b]{0.32\textwidth}
    \centering
    \includegraphics[width=\textwidth]{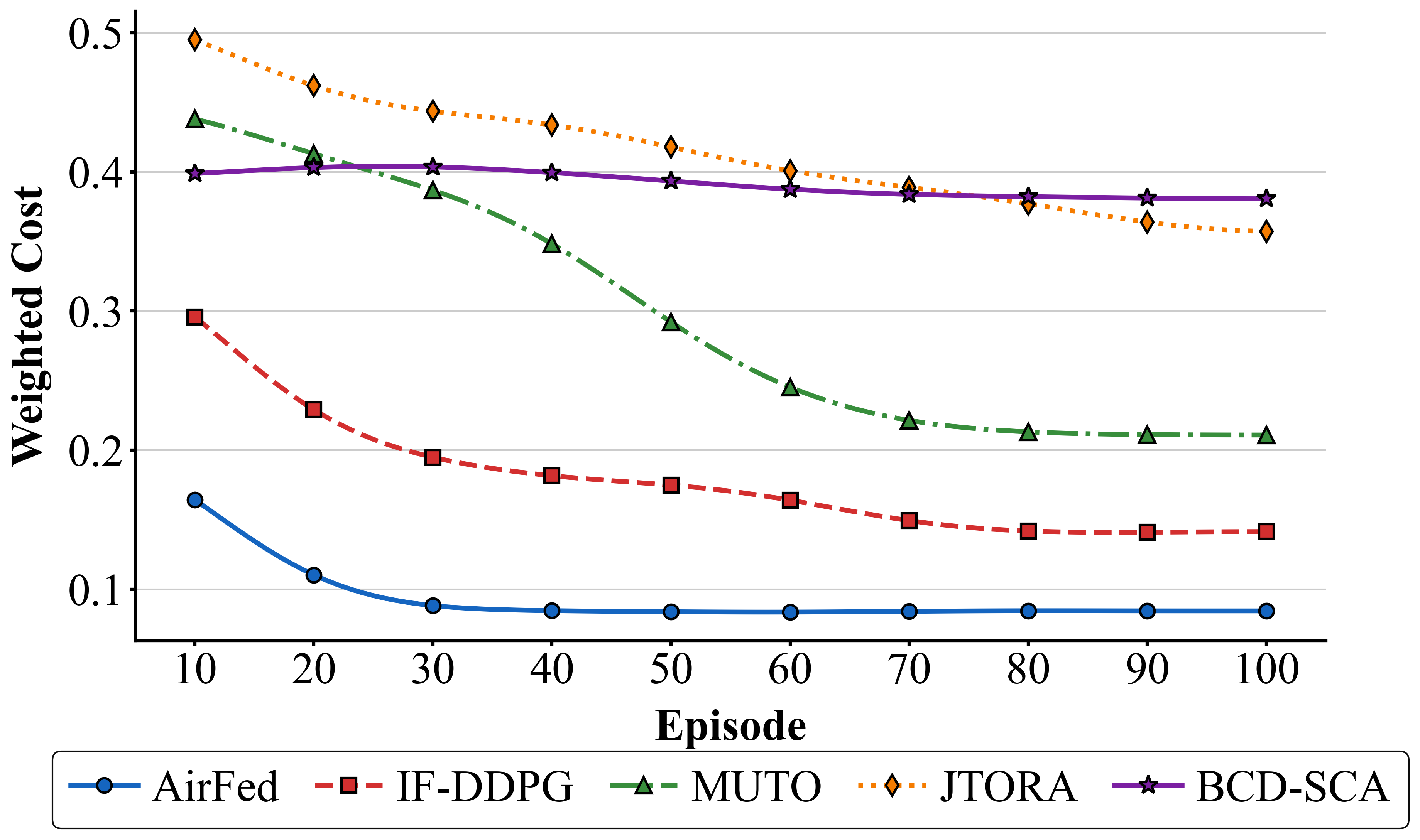}
    \caption{Weighted cost}
    \label{fig:convergence_cost}
\end{subfigure}
\caption{Convergence analysis of AirFed and baseline approaches across key performance metrics.}
\label{fig:convergence}
\end{figure*}

\begin{figure*}[t]
\centering
\begin{subfigure}[b]{0.48\textwidth}
    \centering
    \includegraphics[width=\textwidth]{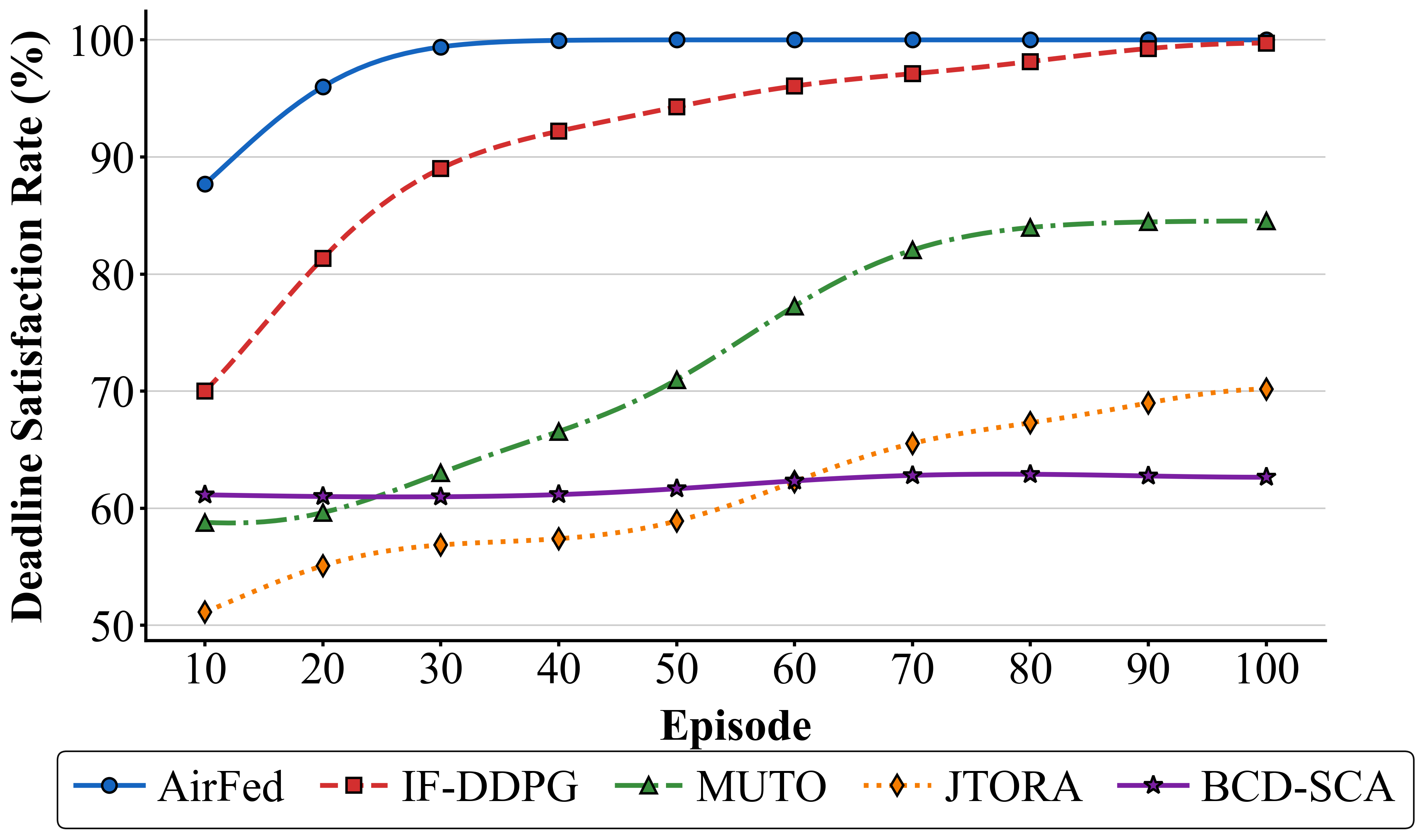}
    \caption{Deadline satisfaction rate}
    \label{fig:qos_deadline}
\end{subfigure}
\hfill
\begin{subfigure}[b]{0.48\textwidth}
    \centering
    \includegraphics[width=\textwidth]{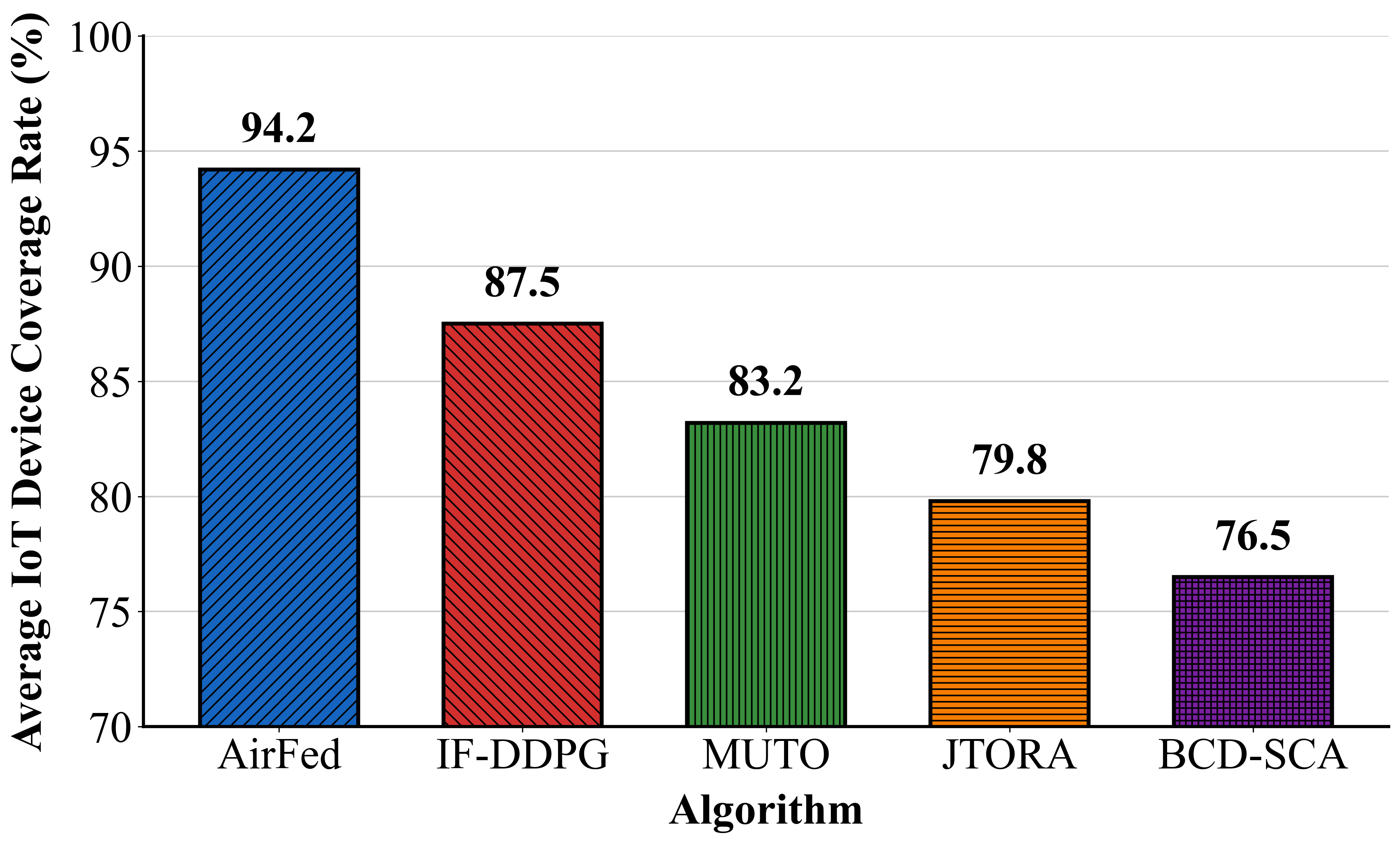}
    \caption{IoT device coverage rate}
    \label{fig:qos_coverage}
\end{subfigure}
\caption{Quality of Service analysis of AirFed and baseline approaches.}
\label{fig:qos}
\end{figure*}

\subsubsection{Hyperparameter Configuration}
The network architecture employs 2-layer GATs (hidden layers [128, 64]) with 4-head attention mechanism, and GRU hidden dimension of 128. In the dual-Actor single-Critic architecture, the shared feature layer has 128 dimensions, two Actors each have 2 layers (hidden layers [128, 128]), and the Critic has 2 layers (hidden layers [128, 64]). Training parameters include discount factor $\gamma = 0.95$, Velocity Actor learning rate $\alpha_{vel} = 3 \times 10^{-4}$, Offloading Actor learning rate $\alpha_{off} = 3 \times 10^{-4}$, Critic learning rate $\alpha_{critic} = 5 \times 10^{-4}$, and entropy regularization coefficient $\beta_{entropy} = 0.01$. Reward function weights $\alpha = \beta = 0.5$ balance time and energy objectives, deadline penalty $\lambda = 10$, coverage reward $\eta = 0.1$. Federated learning parameters include quantization bit widths $b_{min} = 4$ and $b_{max} = 16$, reputation weights $\alpha_{succ} = 0.6$ and $\alpha_{stab} = 0.4$, smoothing factor $\rho = 0.75$. All key hyperparameters are optimized through grid search to ensure optimal performance. Detailed parameters are shown in Table~\ref{tab:hyperparams}.

Baseline approaches are implemented following their original designs. All baselines undergo the same hyperparameter tuning through grid search in our specific experimental setting. For hyperparameters reported in the original papers, we search within ranges around the reported values. For unreported hyperparameters, we conduct broader grid search to find optimal configurations. This ensures fair comparison under identical environmental conditions.

\subsection{Convergence Analysis}
Fig.~\ref{fig:convergence} presents the convergence characteristics of AirFed and baseline approaches on the optimization objective defined in Eq.~\eqref{eq: obj}, evaluated through three key metrics: average task completion time ($F_{time}$), per-UAV energy consumption ($F_{energy}$), and weighted cost ($F_{total}$). AirFed demonstrates superior convergence behavior, achieving stable performance within 30-40 episodes across all metrics. In contrast, IF-DDPG and MUTO require 70-80 episodes to converge, while JTORA exhibits slower convergence, typically stabilizing after 80-100 episodes. As a non-learning approach, BCD-SCA maintains constant performance throughout the training process.

For task completion time (Fig.~\ref{fig:convergence_time}), AirFed converges to around 0.9s, while IF-DDPG, MUTO, and JTORA converge to approximately 1.4s, 3.3s, and 4.7s respectively, and BCD-SCA remains at around 5.0s. Similarly, AirFed converges to around 27.0 kJ for energy consumption (Fig.~\ref{fig:convergence_energy}), compared to approximately 29.0 kJ (IF-DDPG), 32.5 kJ (MUTO), 35.0 kJ (JTORA), and 36.5 kJ (BCD-SCA). For weighted cost (Fig.~\ref{fig:convergence_cost}), AirFed converges to around 0.08, achieving approximately 42.9\%, 61.9\%, 77.8\%, and 80.0\% reduction compared to IF-DDPG (0.14), MUTO (0.21), JTORA (0.36), and BCD-SCA (0.38) respectively.

The rapid convergence of AirFed can be attributed to three key factors: the dual-layer GATs that capture spatial-temporal dependencies for informed decision-making, the dual-Actor architecture that efficiently handles hybrid action spaces, and the reputation-based federated learning that enables knowledge sharing across the UAV network. These mechanisms collectively accelerate the learning process compared to baseline approaches.

\subsection{Quality of Service Analysis}
Fig.~\ref{fig:qos} evaluates the QoS performance through 
deadline satisfaction rate and average IoT device 
coverage rate across episodes, driven by the QoS rewards Eq.~\eqref{eq:ddl_r} and Eq.~\eqref{eq:cov_r}. 

For deadline satisfaction rate (Fig.~\ref{fig:qos_deadline}), AirFed reaches over 99\% satisfaction rate within 30 episodes and maintains stability, while IF-DDPG eventually converges to the same level but requires approximately 90 episodes. MUTO converges to around 85\%, JTORA to around 70\%, and BCD-SCA remains at approximately 63\%. The superior performance of AirFed is attributed to the synergy between deadline penalty terms and federated learning: the penalty drives individual UAVs to learn deadline-aware policies, while federated learning enables rapid propagation of effective strategies across the network. Moreover, the dual-Actor architecture coordinates trajectory and task offloading decisions to satisfy timing constraints through dynamic position adjustment and multi-hop offloading paths.

For IoT device coverage rate (Fig.~\ref{fig:qos_coverage}), AirFed achieves 94.2\%, outperforming IF-DDPG (87.5\%), MUTO (83.2\%), JTORA (79.8\%), and BCD-SCA (76.5\%) by 7.7\%, 13.2\%, 18.0\%, and 23.2\% respectively. This advantage stems from the coverage reward mechanism that incentivizes UAVs to optimize trajectories for broader device coverage, while the service layer in the dual-layer graph structure enables real-time awareness of coverage status.

\subsection{Ablation Analysis}
\begin{figure}[h]
\centering
\includegraphics[width=0.48\textwidth]{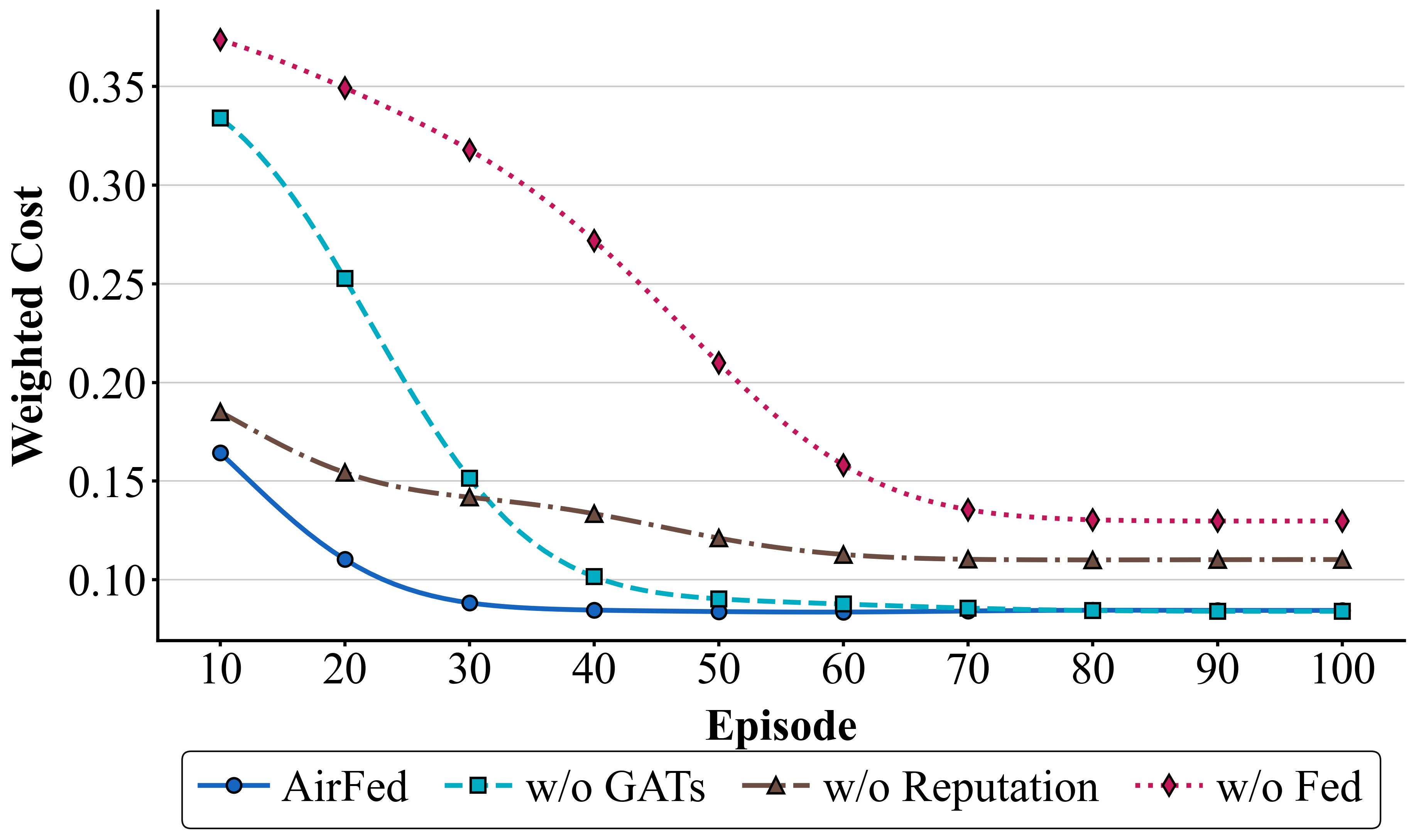}
\caption{Ablation analysis of AirFed key components.}
\label{fig:ablation}
\end{figure}

Fig.~\ref{fig:ablation} illustrates the ablation study results of AirFed's key components. We evaluate three variants: the version without GATs (w/o GATs, replaced by MLP), the version without the reputation mechanism (w/o Reputation, using equal-weight aggregation), and the version without federated learning (w/o Fed, completely independent training).

Removing federated learning (w/o Fed) results in the most significant performance degradation, with the weighted cost increasing from 0.08 to approximately 0.13, representing a 62.5\% increase. This variant requires about 70 episodes to converge, while the complete AirFed only needs 40 episodes, achieving a 42.9\% improvement in convergence speed. Without federated learning, each UAV must independently explore the state space, leading to inefficient learning and difficulty in discovering globally optimal strategies.

Removing the reputation mechanism (w/o Reputation) increases the weighted cost to approximately 0.11, representing a 37.5\% increase compared to the complete version, with convergence time extended to about 60 episodes. Without the reputation mechanism, all UAVs' model parameters are treated equally, causing low-quality or unstable policies to contaminate global knowledge, delaying convergence and reducing final performance. The reputation mechanism ensures that only high-quality policies propagate through the network by filtering unreliable updates.

Removing GATs (w/o GATs) achieves similar final performance to the complete version, but converges significantly slower, requiring about 70 episodes compared to AirFed's 40 episodes. This indicates that GATs play an important role in accelerating the learning process by effectively modeling spatial collaboration relationships among UAVs and dynamic network topology, enabling the system to explore and converge to optimal strategies more quickly. 

\subsection{Communication Efficiency Analysis}
\begin{figure}[t]
\centering
\includegraphics[width=0.48\textwidth]{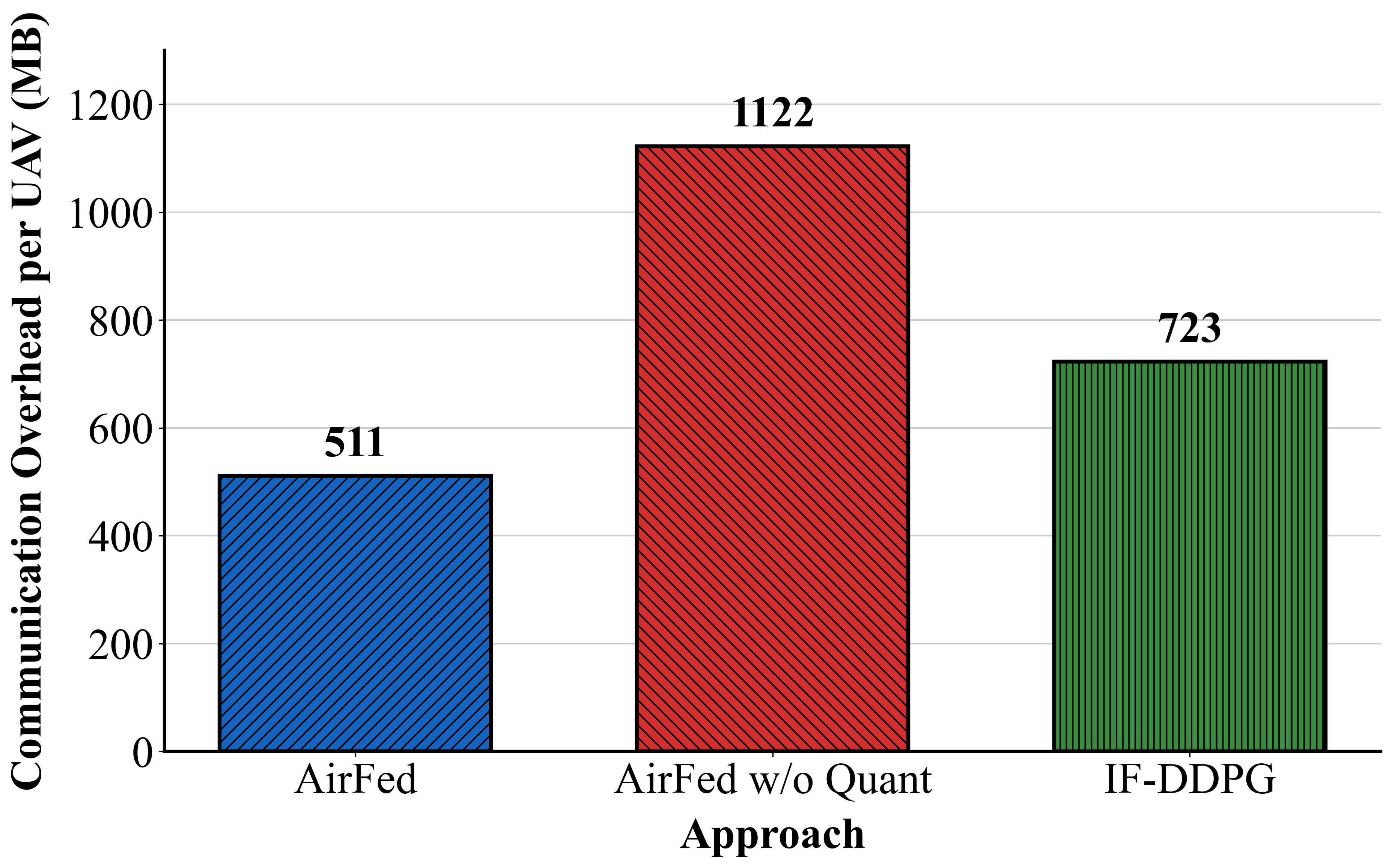}
\caption{Communication overhead comparison of federated learning approaches.}
\label{fig:comm}
\end{figure}

Fig.~\ref{fig:comm} presents the communication overhead comparison between AirFed and two baseline methods. Since MUTO, JTORA, and BCD-SCA do not employ federated learning frameworks and thus do not exchange model parameters among UAVs, they incur no federated learning communication overhead. Therefore, to evaluate the impact of the gradient-sensitive adaptive quantization mechanism, we compare AirFed with AirFed w/o Quant (full-precision 32-bit parameter transmission) and IF-DDPG. AirFed employs gradient-sensitive adaptive quantization with $b_{min}=4$ and $b_{max}=16$ to reduce communication burden in federated learning.

The result shows that during the 100 episodes, AirFed achieves 511 MB communication overhead per UAV, outperforming AirFed w/o Quant (1122 MB) and IF-DDPG (723 MB) by 54.5\% and 29.3\% reduction, respectively. This stems from AirFed's gradient-sensitive adaptive quantization mechanism, which allocates higher bit precision to parameters with large gradients while compressing less sensitive parameters with lower bit widths, reducing bandwidth consumption while preserving critical learning information. This communication efficiency is particularly important for UAV-MEC systems with limited bandwidth and energy budgets.

\begin{figure*}[t]
\centering
\begin{subfigure}[b]{0.32\textwidth}
    \centering
    \includegraphics[width=\linewidth]{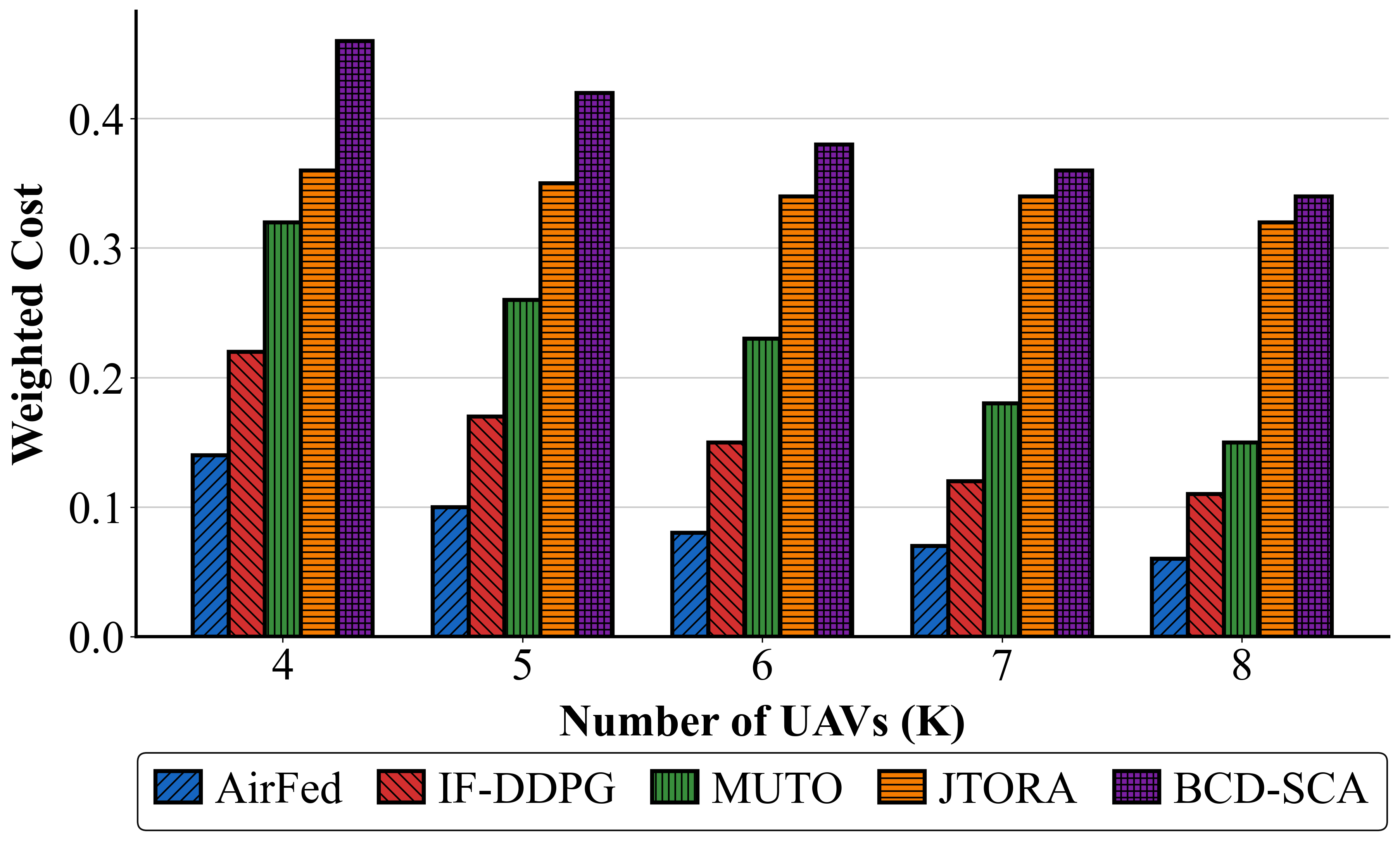}
    \caption{Varying UAV numbers}
    \label{fig:scale_uav}
\end{subfigure}
\hfill
\begin{subfigure}[b]{0.32\textwidth}
    \centering
    \includegraphics[width=\linewidth]{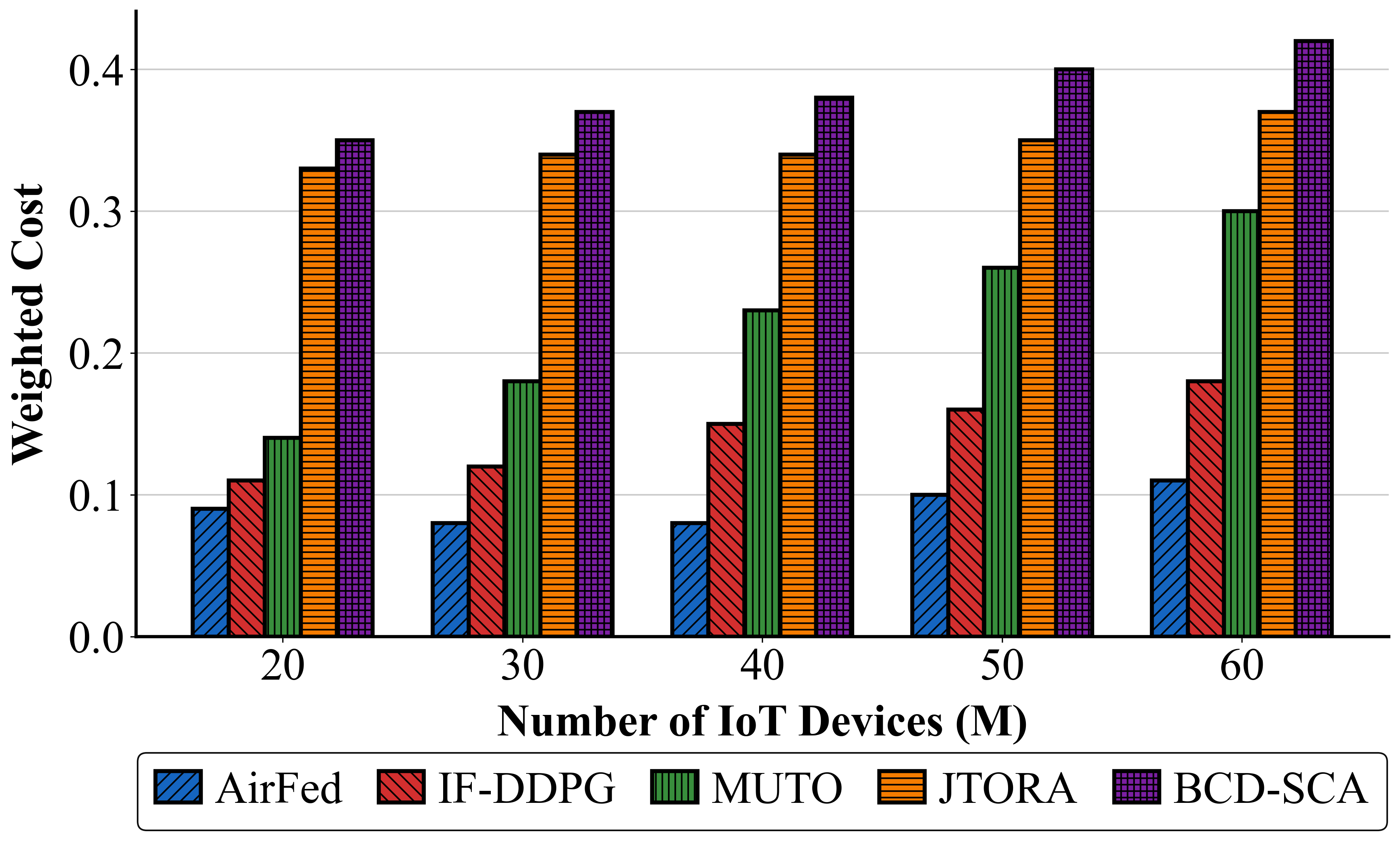}
    \caption{Varying IoT device numbers}
    \label{fig:scale_iot}
\end{subfigure}
\hfill
\begin{subfigure}[b]{0.32\textwidth}
    \centering
    \includegraphics[width=\linewidth]{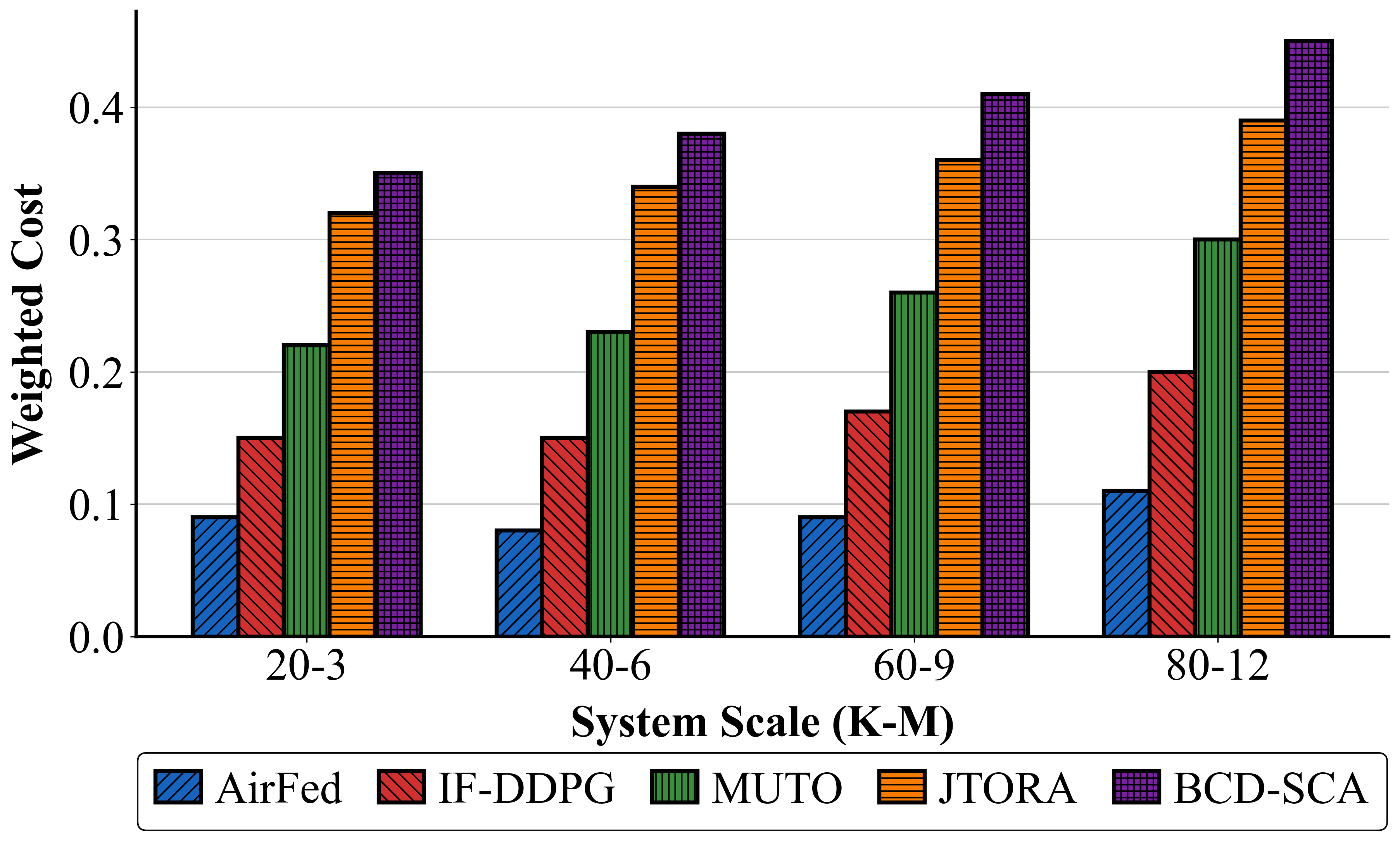}
    \caption{Varying system scale}
    \label{fig:scale_system}
\end{subfigure}
\caption{Scalability analysis of AirFed across UAV numbers, IoT device numbers, and system scale.}
\label{fig:scalability}
\end{figure*}

\subsection{Scalability Analysis}
To evaluate AirFed's scalability, we analyze its performance across three dimensions: UAV numbers, IoT device numbers, and system scale.

Fig.~\ref{fig:scale_uav} shows the performance variation when UAV numbers increase from 4 to 8 with fixed IoT devices (M=40). The weighted cost of all approaches decreases as UAV numbers increase. At the 8-UAV configuration, AirFed achieves around 0.06, outperforming IF-DDPG (0.11), MUTO (0.15), JTORA (0.32), and BCD-SCA (0.34) by 45.5\%, 60.0\%, 81.3\%, and 82.4\%, respectively. Fig.~\ref{fig:scale_iot} shows the impact when IoT devices increase from 20 to 60 with fixed UAVs (K=6). As device numbers increase, the weighted cost of all approaches rises. AirFed exhibits the smallest increase (from 0.09 to 0.11), and at the M=60 configuration outperforms IF-DDPG (0.18), MUTO (0.30), JTORA (0.37), and BCD-SCA (0.42) by by 38.9\%, 63.3\%, 70.3\%, and 73.8\%, respectively. Fig.~\ref{fig:scale_system} shows the performance when system scale expands from 20-3 to 80-12 (maintaining K/M$\approx$6.7). The weighted cost of all approaches increases with scale growth. At the 80-12 configuration, AirFed achieves 0.11, outperforming IF-DDPG (0.20), MUTO (0.30), JTORA (0.39), and BCD-SCA (0.45) by 45.0\%, 63.3\%, 71.8\%, and 75.6\%, respectively.

The superior scalability of AirFed across all three dimensions is attributed to several key architectural features. First, the decentralized federated learning mechanism fully exploits the collaborative potential of additional UAVs, with denser knowledge sharing networks accelerating policy propagation and global convergence as UAV numbers increase. Second, the coverage-aware reward mechanism drives UAVs to proactively adjust trajectories for broader device coverage, while the dual-layer GATs effectively model complex spatial relationships arising from increased device density, maintaining robustness as IoT device numbers grow. Finally, the distributed decision-making architecture enables parallel processing across UAVs, and maintains computational efficiency as system scale expands. In contrast, baseline approaches suffer from performance saturation or significant degradation as system scale increases due to a lack of efficient collaboration mechanisms.

\section{Conclusions and Future Work}
\label{conclusions}
AirFed is a federated graph-enhanced multi-agent reinforcement learning framework for multi-UAV cooperative mobile edge computing. It models spatial-temporal dependencies through a dual-layer dynamic graph attention network, handles hybrid action spaces via a dual-Actor single-Critic architecture, and achieves efficient knowledge sharing through reputation-based decentralized federated learning. Experimental results demonstrate that the AirFed framework reduces weighted cost by 42.9\% compared to the best baseline, converges within 30-40 episodes, and achieves over 99\% deadline satisfaction rate and 94.2\% IoT device coverage rate. Ablation, communication efficiency and scalability analysis demonstrate the framework's effectiveness of each component and robust performance across different system scales.

Future work includes integrating privacy-preserving mechanisms (e.g., post-quantum cryptography) to safeguard data security during federated learning, investigating robust learning strategies under adversarial conditions including communication interference and UAV failures, and conducting prototype validation and field testing on real UAV platforms to evaluate practical deployment performance.

\ifCLASSOPTIONcaptionsoff
  \newpage
\fi

\bibliographystyle{IEEEtran}
\bibliography{IEEEabrv,Bibliography}
\vspace{-6mm}

\vfill

\end{document}